\documentclass{simulacrum}
\usepackage{auxiliary/neurips_packages}
\usepackage{auxiliary/mymacros}
\usepackage{auxiliary/filecommands}
\usepackage{adjustbox}
\usepackage{longtable}
\usepackage{xurl} %

\providecommand{\answerYes}[1][]{\textcolor{blue}{[Yes]#1}}

\providecommand{\answerNA}[1][]{\textcolor{gray}{[N/A]#1}}
\providecommand{\answerTODO}[1][]{\textcolor{red}{\bf [TODO]}}
\providecommand{\justificationTODO}[1][]{\textcolor{red}{\bf [TODO]}}

\title[TempusBench]{\tempusbench{}: An Evaluation Framework for Time-Series Forecasting}
\author{Denizalp Goktas\affil{*} \and Gerardo Ria\~no-Brice\~no\affil{*} \and Omkar Tekawade\affil{*} \and Alif Abdullah \and Md Yasif Jamal \and Vinesha Shaik \and Arsh Singh \and Aryan Nair \and Chenkai Shen \and Beatriz de Lucio \and Alexandra Magnusson \and Farhan Mashrur \and Ahmed Abdulla \and Shawrna Sen \and Mahitha Thippireddy \and Gregory Schwartz \and Amy Greenwald}
\affiliations{Simulacrum, New York City, NY, USA \quad \affil{*}Equal contribution.}
\correspondence{\href{mailto:deni@smlcrm.com}{deni@smlcrm.com}, \href{mailto:amy@smlcrm.com}{amy@smlcrm.com}}
\reporttype{Technical Report}

\begin{document}

\maketitle
\begin{abstract}
    Foundation models have transformed natural language processing and computer vision, and a rapidly growing literature on time-series foundation models (TSFMs) seeks to replicate this success in forecasting. While recent open-source models demonstrate the promise of TSFMs, the field lacks a comprehensive and community-accepted model evaluation framework. We see at least four major issues impeding progress on the development of such a framework. First, existing evaluation frameworks comprise benchmark forecasting tasks derived from often outdated datasets (e.g., M3), many of which lack clear metadata and overlap with the corpora used to pre-train TSFMs. %
    Second, these frameworks evaluate models along a narrowly defined set of benchmark forecasting tasks, such as forecast horizon length or domain, but overlook core statistical properties such as non-stationarity and seasonality. Third, domain-specific models (e.g., XGBoost) are often compared unfairly, as existing frameworks do not enforce a systematic and consistent hyperparameter tuning convention for all models. Fourth, visualization tools for interpreting comparative performance are lacking.
    To address these issues, we introduce \tempusbench{}, an open-source evaluation framework for TSFMs. \tempusbench{} consists of 1) new datasets which are not included in existing TSFM pretraining corpora, 2) a set of novel benchmark tasks that go beyond existing ones, 3) a model evaluation pipeline with a standardized hyperparameter tuning protocol, and 4) a tensorboard-based visualization interface. We provide access to our code on GitHub: \repo{} and maintain a live leaderboard at \dashboard{}.
\end{abstract}

\section{Introduction}

Foundation models are models trained on large and diverse datasets that can be used to solve a variety of downstream tasks. Their success in natural language processing (NLP) and computer vision has inspired an emerging literature on \mydef{time-series foundation models (TSFMs)}. (TSFMs) are models that take past time-series data (and possibly covariate time-series data) as input, and output future values (or distributions over them), typically formulated as neural networks trained via supervised learning. 
While about a dozen open-source TSFMs \cite{woo2024moirai, liu2024moirai,rasul2023lag, ansari2024chronos,das2023timesfm,goswami2024moment,auer2025tirex, hemmer2025true,ansari2025chronos,hoo2025tables, feng2025kairos,ekambaram2024tiny,cohen2025time,liu2025sundial}
are now available, comparing their performance to one another and to traditional domain-specific models (e.g., ARIMA \cite{box1970time}, SVR \cite{vapnik1995nature, drucker1997support}) remains difficult. A handful of evaluation frameworks \cite{godahewa2021monash,qiu2024tfb, zeng2023transformers, shao2024exploring, zhang2024probts, aksu2024gift} have been released, but the field still lacks comprehensive, community-accepted standards for model evaluation \cite{liang2024tutorial}, which creates an impediment to replicating the success that foundation models have seen in NLP and computer vision \citep{kalekar2004time}.

In our view, existing evaluation frameworks face four major challenges.
First, the evaluation ecosystem relies on outdated datasets such as M3 \cite{makridakis2000m3} and M4 \cite{makridakis2018m4}, many lacking metadata (e.g., variable names). More importantly, the existing evaluation datasets overlap with the pretraining corpora of TSFMs, leading to inflated estimates of zero-shot generalization \citep{anand2024comparative}. For example, except for Moirai2, all TSFMs assessed by GIFT-Eval include test data in their training corpora \cite{aksu2024gift, salesforce2024giftevalapp}.
Second, current frameworks define benchmark forecasting tasks only along narrow axes (i.e., forecast horizon, variate type, frequency, and domain). While useful, these dimensions neglect key statistical properties long studied in time-series analysis such as (non-)stationarity and seasonality. Since they do not evaluate such properties, existing frameworks cannot thoroughly evaluate model capabilities.
Third, existing frameworks have not yet developed standardized hyperparameter tuning routines, leading to unfair comparisons between TSFMs and domain-specific models, as the performance of domain-specific models depends heavily on hyperparameter choices.%
\footnote{TSFMs require hyperparameter searches during pretraining, but not during evaluation.} 
Indeed, as noted by practitioners \cite{redditTSMFs}, simple statistical models with well-chosen hyperparameters can outperform more complex ones, highlighting the need for consistent tuning routines.
Fourth, evaluation typically reduces to numerical metrics such as mean squared error, which practitioners remark \cite{hyndman2006another} provide limited interpretability. For instance, under GIFT-Eval, Seasonal Naive \cite{hyndman2021fpp3, makridakis2000m3} outperforms five open-source TSFMs, but this empirical observation offers no insight into the strength and weaknesses of TSFMs, since seasonal naive fails when seasonality is weak. Beyond quantitative scores, qualitative analyses---especially forecast visualizations---are essential.

To address these issues, we introduce \tempusbench{}, an open-source evaluation framework. \tempusbench{} consists of 1) new datasets which are not included in existing TSFM pretraining corpora, 2) a set of novel benchmark tasks that goes beyond existing ones, and 3) a model evaluation pipeline with a standardized hyperparameter tuning protocol, and 4) a tensorboard-based visualization interface.

\makeatletter\@ifclassloaded{simulacrum}{\par}{\vspace{-0.66em}}\makeatother
\paragraph{Contributions}
\tempusbench{} includes 53 forecasting models,%
\footnote{In \Cref{sec_app:models}, we list the families of forecasters in 
TempusBench (\Cref{tab:models_summary}) and describe many of them. 47 of these models are multivariate forecasters with covariate support, while 50 are multivariate forecasters without covariate support.}
a number of which (e.g., XGBoost) have not previously been considered by evaluation frameworks.
Moreover, it is designed to overcome the aforementioned four issues by improving along the following dimensions.
First, we introduce 70 new time-series datasets that are not based on existing time-series evaluation datasets, and that are not contained in the training corpus of open-source TSFMs released to date. 
Second, we propose new benchmark task types that extend beyond horizon length, variate type, frequency, and domain. These include categories based on stationarity, seasonality, variable type (continuous, count, binary, categorical), dataset size (small vs.\ large), sparsity (sparse vs.\ dense), and quality (noisy vs.\ measurement error).
Third, we introduce a model evaluation pipeline that runs a standardized and automated hyperparameter search procedure for all forecasting models with hyperparameters, facilitating a fair comparison of all forecasting methods.
Fourth, \tempusbench{} comes packaged with a tensorboard-based visualization application that easily allows researchers and practitioners to visualize and interpret the performance of different models on different tasks.

\begin{table*}[!htbp]
\centering
\small
\caption{Property comparisons of various forecasting benchmarks.}
\label{tab:benchmark_comparison_transposed}
\setlength{\tabcolsep}{0.5pt}
\renewcommand{\arraystretch}{1.2}
\begin{tabular}{@{}l c c c c c c c@{}}
\hline
{Property} & 
{Monash \cite{godahewa2021monash}} & 
{TFB \cite{qiu2024tfb} } & 
{LTSF \cite{zeng2023transformers}} & 
{BasicTS+ \cite{shao2024exploring}} & 
{ProbTS \cite{zhang2024probts}} & 
{GIFT-Eval \cite{aksu2024gift}} &
TempusBench\\
\hline \hline
\makecell[l]{Frequency \\ Range} & \makecell{Second \\to  Year} & \makecell{Minute \\to  Year}  & \makecell{Minute \\to  Week} & \makecell{Minute \\to  Day} & \makecell{Minute \\to  Week} & \makecell{Second \\to Year} & \makecell{Second \\to Year}  \\ \hline
\makecell[l]{Num.\\  Domains}  & 7 & 6 & 5 & 3 & 5 & 7 & 10 \\ \hline
\makecell[l]{Train/Test\\ data leak} & \redcmark & \redcmark & \redcmark & \redcmark & \redcmark & \redcmark & \greenxmark \\ \hline
\makecell[l]{Variate\\ Types} & Uni & Uni/Multi & Multi & Multi & Multi & Uni/Multi & Uni/Multi \\ \hline
\makecell[l]{Prediction \\ Length} & Short & Short & Long & Short/Long & Short/Long & Short/Long & Short/Long \\ \hline
\makecell[l]{Stat. Benchmarks} & \redxmark & \redxmark & \redxmark & \redxmark & \redxmark & \redxmark & \greencmark \\ \hline
\makecell[l]{Forecaster types} & Stat./DL & Stat./DL & Stat./DL & Stat./DL & Stat./DL/FM & Stat./DL/FM & Stat./ML/DL/FM \\ \hline
\makecell[l]{Hyperparam.\\ autotuning} & \redxmark & \redxmark & \redxmark & \redxmark & \redxmark & \redxmark & \greencmark \\
\hline
\vspace{-1cm}
\end{tabular}
\end{table*}

\section{Background}

We refer the reader to \Cref{sec_app:prelims} for the notational convention we adopt, and for additional mathematical preliminaries and evaluation metric definitions.

\paragraph{Forecasting Tasks}

A \mydef{(time-series) forecasting task} $\forecastprob \doteq (\contextlen, \forecastlen, \numcovars, \numtargets, (\covarset[\covar])_{\covar \in [\numcovars]}, (\targetset[\target])_{\target \in [\numtargets]}, \covarts, \targetts)$ comprises a \mydef{context length} $\contextlen \in \N$, a \mydef{forecast horizon} $\forecastlen \in \N$, 
$\numtargets \in \N$ \mydef{target time-series} $\targetts = (\targetts[1], \hdots, \targetts[\numtargets])^T$,
where for each variate $\target \in \targets$, entries of $\targetts[\target] \in \targetset[\target]^{\contextlen}$ take values $\targetts[\target][t]$, for $t \in [\contextlen]$, from a \mydef{set of target values} %
$\targetset[\target] \subseteq \R$,
and $\numcovars \in \N$ \mydef{covariate time-series} $\covarts = (\covarts[1], \hdots, \covarts[\numcovars])^T$,
where for each covariate $\covar \in \covars$, $\covarts[\covar] \in \covarset[\covar]^{\contextlen + \forecastlen}$ takes values $\covarts[\covar][t]$, for $t \in [\contextlen + \forecastlen]$, from a \mydef{set of covariate values} %
$\covarset[\covar] \subseteq \R$.
We denote the \mydef{joint set of covariate values} by $\covarset^{\contextlen + \forecastlen} \doteq \bigtimes_{\covar \in \covars} \covarset[\covar]^{\contextlen + \forecastlen}$ and the \mydef{joint set of target variate values} by $\targetset^{\contextlen} \doteq \bigtimes_{\target \in \targets} \targetset[\target]^{\contextlen}$.

A forecasting task $\forecastprob$ is said to be \mydef{univariate} iff $\numtargets = 1$; if $\numtargets > 1$, it is said to be \mydef{multivariate}. A \mydef{covariate} forecasting task $\forecastprob$ is one for which $\numcovars > 0$. 
A forecasting task $\forecastprob$ is said to be continuous (resp.\ count | categorical | binary) iff for all $\target \in \targets$ $\targetset[\target] \subseteq \R$ is a continuous set (resp.\ $\targetset[\target] = \N$ | $\targetset[\target] \subsetneq \N$ | $\targetset[\target] = \{0, 1\}$).

\paragraph{Forecasters}

Given a forecasting task $\forecastprob$, a \mydef{(point) forecast} is a matrix $\forecastts \doteq (\forecastts[1], \hdots, \forecastts[\numtargets])^T$ s.t.\ for all target variates $\target \in \targets$, $\forecastts[\target] \in \targetset[\target]^{\forecastlen}$
represents forecasted values of variate $\target$ for $\forecastlen$ time steps. 
A \mydef{(point) forecasting model} 
is a mapping $\forecaster: \covarset^{\contextlen + \forecastlen} \times \targetset^{\contextlen} \to \targetset^{\forecastlen}$ whose output is a forecast for $\forecastprob$: i.e., $\forecaster(\covarts, \targetts) \doteq (\forecaster[1](\covarts, \targetts), \hdots, \forecaster[\numtargets](\covarts, \targetts))^T = \forecastts$.
More generally, a \mydef{probabilistic forecasting model} is a mapping $\stochforecaster: \covarset^{\contextlen + \forecastlen} \times \targetset^{\contextlen} \to \simplex(\targetset^{\forecastlen})$ s.t. $\stochforecaster(\covarts, \targetts)[\forecastts]$ denotes the realization probability of $\forecastts \in \targetset^{\forecastlen}$.

\paragraph{Forecasting Evaluation Frameworks}
In practice, many forecasters $\forecaster[][\param]: \covarset^{\contextlen + \forecastlen} \times \targetset^{\contextlen} \to \targetset^{\forecastlen}$ depend on hyperparameters $\param \in \params$, so that it is often more appropriate to consider a \mydef{family of forecasters} $\forecasters[][\params] \doteq \{ \forecaster[][\param] \}_{\param \in \params}$, and then choose the forecaster with parameters best suited to the forecasting task at hand.\footnote{For instance, the forecast of an ARIMA model is dependent on choices of hyperparameters given by the order of the number of time lags, the degree of differencing, and the order of the moving-average model. It thus feels more appropriate to talk about the family of ARIMA models.}
A forecaster evaluation framework $\evalframework \doteq (\nummodels, \numbenchmarks, \evaluator, \{\params[\model]\}_{\model =1}^{\nummodels}, \{\forecasters[][{\params[\model]}]\}_{\model =1}^{\nummodels}, \{\forecastprob[\benchmark] \}_{\benchmark \in \benchmarks})$ comprises a family of $\{ \forecasters[][{\params[\model]}] \}_{\model \in \modelset}$ of $\nummodels \in \N$ forecasters, each one defined by a set of hyperparameters $\params[\model]$; $\numbenchmarks \in \N$ forecasting tasks (or, colloquially, \mydef{benchmark tasks} or just \mydef{benchmarks}) $\{\forecastprob[\benchmark] \}_{\benchmark \in \benchmarks}$;
and a hyperparameter tuner $\evaluator$, which takes as input a benchmark and suitable hyperparameters.

\section{\tempusbench{}}

TempusBench, denoted $\evalframework[{\mathrm{TB}}]$, is a forecasting evaluation framework with a hyperparameter tuner $\evaluator[\mathrm{TB}]$ that optimizes using the following time-series cross validation procedure: 
given a benchmark task, a set of hyperparameters, and a family of forecasters, 
1) rolling evaluation windows are generated by sliding a fixed-size window along the target time series, where each window is partitioned into context, training, and validation segments, and the stride is set equal to the length of the validation segment; 
2) for all hyperparameter combinations, all models are fit, 
given available past data (context + training data), and evaluated on the corresponding validation segments; the hyperparameters selected are those corresponding to the lowest validation loss;%
\footnote{In TempusBench, we use MAE as the validation loss function for optimizing hyperparameters.} 
3) the test loss for each metric (e.g., MSE, RMSE, etc.) is then computed 
using the hyperparameters selected,
and these losses are averaged across all windows to obtain the final metric values. 
This rolling window approach avoids the look-ahead bias that would arise from standard $k$-fold cross validation, and instead respects temporal dependencies in time-series data by maintaining strict temporal ordering.

\begin{table*}[t]
\centering
\small
\setlength{\tabcolsep}{6pt}
\renewcommand{\arraystretch}{1.15}
\caption{Taxonomy of all univariate, multivariate, and covariate tasks included in \tempusbench{}.}
\label{tab:task_categories}
\begin{tabularx}{\linewidth}{@{}lX@{}}
\toprule
\textbf{Category} & \textbf{Benchmark Tasks}\\
\midrule
Movement & Stationary, Non-Stationary\\
Data Quality & Noisy data, Data with measurement error\\
Frequency & Seconds, Minutes, Hours, Days, Weeks, Months, Quarterly, Years\\
Context Length  & 12, 18, 32, 45, 64, 128, 200, 256, 450, 512, 1024, 1536, 2048\\
Forecast Horizon & 4, 5, 6, 8, 16, 17, 22, 30, 33, 35, 39, 42, 55, 64\\
Seasonality & Cyclical, Non-Stationary cyclical, Regressive, Irregular, Additive, Multiplicative\\
Domain  & Energy, Transport, Climate, Software, Web, Sales, Nature, Econ., Healthcare, Manufacturing\\
Dataset Coverage & sparse, dense\\
Target Type & continuous, count, binary, categorical\\
\bottomrule
\end{tabularx}
\end{table*}

In Table~\ref{tab:task_categories}, we present a taxonomy of the benchmarks included in TempusBench.
We summarize the actual benchmarks in Tables~\ref{tab:benchmark_summary_uni}, \ref{tab:benchmark_summary_multi}, and \ref{tab:benchmark_summary_co} (\Cref{sec_app:benchmarks}).

\section{Result Aggregation Procedure}
\label{sec:aggregation}

After evaluating multiple forecasting models across a diverse set of benchmark tasks, we require aggregation methods to summarize and compare model performance at the aggregate level. This section describes two complementary aggregation procedures: \emph{average win rate} and \emph{skill score}.

In practice, some models may not produce valid results on certain tasks (e.g., due to computational failures or data incompatibilities), resulting in missing values. Our aggregation procedures handle these missing values gracefully by excluding unavailable comparisons.

\paragraph{Problem Setup}

Let $\nummodels$ denote the number of models under evaluation and let $\numbenchmarks$ denote the number of benchmark tasks. 
We compute an evaluation (i.e., error) metric $\metric[\model,\benchmark]$ (e.g., MAE, RMSE, etc.) for all models $\model \in \modelset$ on all tasks $\benchmark \in \benchmarks$, and then aggregate those errors in two ways.

\paragraph{Average Win Rate}

The \emph{average win rate} $W_{\model}$ for model $\model$ quantifies the probability that model $\model$ achieves lower error than another randomly chosen model $\model' \neq \model$ on a randomly chosen benchmark task. This metric provides a pairwise comparison perspective that is robust to the absolute scale of errors across different tasks.

Formally, for model $\model$, the average win rate is computed as:
\begin{equation}
W_{\model} = \frac{1}{|\mathcal{C}_{\model}|} \sum_{\benchmark \in \benchmarks} \sum_{\substack{\model' \in \modelset\\ \model' \neq \model}} w_{\model,\model',\benchmark}
\end{equation}

The win indicator $w_{\model,\model',\benchmark}$ is defined as follows:
\begin{equation}
w_{\model,\model',\benchmark} = \begin{cases}
1 & \text{if } \metric[\model,\benchmark] < \metric[\model',\benchmark] \text{ and both values are valid} \\
0.5 & \text{if } \metric[\model,\benchmark] = \metric[\model',\benchmark] \text{ and both values are valid} \\
0 & \text{if } \metric[\model,\benchmark] > \metric[\model',\benchmark] \text{ and both values are valid} \\
0 & \text{if either value is missing}.
\end{cases}
\end{equation}

The normalization factor $|\mathcal{C}_{\model}|$ accounts for the actual number of valid comparisons:
\begin{equation}
\mathcal{C}_{\model} = \{ \benchmark \in \benchmarks, \model' \in \modelset, \model' \neq \model \mid \metric[\model,\benchmark] \textrm{ and } \metric[\model',\benchmark] \textrm{ are both valid} \}
\end{equation}

\paragraph{Skill Score}

The \emph{skill score} $S_{\model}$ for model $\model$ quantifies how much the model reduces forecasting error compared to a fixed baseline model $\beta \ne \model$. Unlike win rate, which compares models in a pairwise manner, skill score provides an absolute measure of improvement relative to a reference model (typically a simple baseline, such as Seasonal Naive Forecasting).

For model $\model$ relative to baseline $\beta$, the skill score is computed as:
\begin{equation}
S_{\model} = 1 - \left(\prod_{\benchmark \in \mathcal{R}_{\model}} \mathrm{clip}\left(\frac{\metric[\model,\benchmark]}{\metric[\beta,\benchmark]}; \ell, u\right)\right)^{1/|\mathcal{R}_{\model}|},
\end{equation}
where $\mathcal{R}_{\model} = \{\benchmark \in \benchmarks : \metric[\model,\benchmark] \text{ and } \metric[\beta,\benchmark] \text{ are both valid}\}$ is the set of tasks for which both model $\model$ and baseline $\beta$ have valid results, and $\mathrm{clip}(x; \ell, u) = \max(\ell, \min(x, u))$ clips the relative error ratio to the interval $[\ell, u]$ with $\ell = 10^{-2}$ and $u = 100$.

The clipping operation prevents extreme relative errors (e.g., division by near-zero baseline errors) from dominating the geometric mean. 
We handle the edge case $\metric[\beta,\benchmark] = 0$ as follows:
\begin{equation}
\frac{\metric[\model,\benchmark]}{\metric[\beta,\benchmark]} = \begin{cases}
1 & \text{if } \metric[\model,\benchmark] = 0, \\
u & \text{if } \metric[\model,\benchmark] > 0.
\end{cases}
\end{equation}

The skill score interpretation is straightforward:
\begin{itemize}
\item $S_{\model} > 0$: Model $\model$ performs better than the baseline (lower relative error).
\item $S_{\model} = 0$: Model $\model$ performs equivalently to the baseline.
\item $S_{\model} < 0$: Model $\model$ performs worse than the baseline.
\end{itemize}

\paragraph{Geometric Mean Rationale}

The skill score uses a geometric mean (via the product raised to the reciprocal power) rather than an arithmetic mean for aggregating relative errors across tasks. This choice has several advantages:
1.~\emph{Scale invariance}: The geometric mean is invariant to multiplicative scaling, so that high-error tasks do not dominate. Tasks with different error magnitudes contribute proportionally.
2.~\emph{Symmetry}: The geometric mean treats improvements and degradations symmetrically (e.g., a 2$\times$ improvement and a 2$\times$ degradation cancel out in the geometric mean).
3.~\emph{Robustness}: The geometric mean is less sensitive to outliers than the arithmetic mean, which is important when aggregating across diverse benchmark tasks.

\paragraph{Implementation Details}

Both aggregation procedures are available in \href{https://github.com/Smlcrm/TempusBench}{TempusBench}, which
provides two aggregator classes: \texttt{WinRate} and \texttt{SkillScore}, both inheriting from \texttt{BaseAggregator}. 
Each can be instantiated with a pivot table, where rows represent models $\model \in \modelset$, columns represent benchmark tasks $\benchmark \in \benchmarks$, and cell $(i,j)$ stores error $\metric[\model,\benchmark]$; and, in the case of \texttt{SkillScore}, a baseline model $\beta$ (default: $\beta = \texttt{seasonal\_naive}$).
Missing values are automatically detected and excluded from the aggregation, ensuring that models are only compared on tasks where both models have valid results.

\subsection{Average Win Rate Results}

In Tables~\ref{tab:winrate_univariate}, \ref{tab:winrate_multivariate}, and~\ref{tab:winrate_covariate}, we report the per-metric average win rates for deterministic and probabilistic models, across univariate, multivariate, and covariate forecasting tasks, respectively. All win rates compare a model's performance pairwise against competing models on the evaluated tasks. Higher win rates indicate that a model consistently outperforms its pairwise competitors, while lower win rates indicate more frequent underperformance.

\paragraph{Univariate (see Table~\ref{tab:winrate_univariate})}
The strongest models (TiRex, TiRex 1.1, TimesFM 2.5, Moirai 2.0, and Chronos-2) achieve roughly 64\%–75\% average win rates across both stochastic (CRPS, QS, WIS) and deterministic (MAE, MAPE, MASE, RMSE) metrics—indicating consistent superiority across task pairings. TiRex leads across all measured metrics, peaking at over 75\% win rates for MAE and MASE, closely followed by its 1.1 variant and TimesFM 2.5. Chronos-2 and Moirai 2.0 also show robust performance, winning in about two-thirds of the evaluated tasks. Lighter models, such as the Chronos-Bolt series and PatchTST-FM, remain highly competitive with win rates generally between 55\% and 66\%, illustrating solid baseline capabilities even at smaller parameter counts.

\paragraph{Multivariate (see Table~\ref{tab:winrate_multivariate})}
TiRex and TiRex 1.1 establish a dominant lead on multivariate tasks, attaining win rates between 78\% on stochastic metrics and over 85\% on deterministic metrics (MAE, MASE, RMSE). TimesFM 2.5 and Chronos-2 maintain strong second-tier performance, winning over 73\% of their respective evaluations. Notably, Chronos-2 exhibits heterogeneous metric behavior: high win rates on CRPS, QS, WIS, MAE, MASE, and RMSE (77\%–83\%), but a noticeable drop on MAPE (64.5\%). Models like PatchTST-FM and TabPFN-TS show highly competitive deterministic scores (particularly MAE and MASE around 72\%–76\%) despite lower comparative stochastic performance.

\paragraph{Covariate (see Table~\ref{tab:winrate_covariate})}
Chronos-2 and TabPFN-TS demonstrate the strongest performance in covariate forecasting. Chronos-2 dominates the probabilistic metrics with an 83\% win rate across CRPS, QS, and WIS, while TabPFN-TS commands the deterministic metrics, peaking at 82\% on MAE and MASE, and 80.4\% on MAPE. Chronos-2-Small similarly shows robust performance, outperforming many significantly larger models. Notably, TiRex and TiRex 1.1, which dominated univariate and multivariate tasks, drop to the 10th and 11th ranks here, with win rates hovering around 60\%–67\%. Several classical and statistical baselines, such as Prophet, perform surprisingly well on deterministic metrics (e.g., nearly 70\% win rate on RMSE).

\begin{table}[tpb]
\centering
\caption{\label{tab:winrate_univariate}
Univariate tasks -- Average win rates by metric for deterministic \& probabilistic models}
\resizebox{1\textwidth}{!}{%
\begin{tabular}{r|l|c|ccc|cccc}
\toprule
  \multirow{2}{*}{Rank} & \multirow{2}{*}{Model} & \multirow{2}{*}{Params (M)} & \multicolumn{3}{c|}{Stochastic} & \multicolumn{4}{c}{Deterministic} \\
  & & & CRPS & QS & WIS & MAE & MAPE & MASE & RMSE \\
\midrule
  1 & TiRex (NX-AI) & 300 & \cellcolor{green!15}\textbf{70.4\%} & \cellcolor{green!15}\textbf{70.4\%} & \cellcolor{green!15}\textbf{70.4\%} & \cellcolor{green!15}\textbf{75.1\%} & \cellcolor{green!15}\textbf{70.5\%} & \cellcolor{green!15}\textbf{75.1\%} & \cellcolor{green!15}\textbf{73.4\%} \\
  2 & TiRex 1.1 (NX-AI) & 300 & 68.5\% & 68.5\% & 68.5\% & 74.3\% & 69.7\% & 74.3\% & 71.5\% \\
  3 & TimesFM 2.5 (Google) & 200 & 66.4\% & 66.4\% & 66.4\% & 73.3\% & 69.4\% & 73.4\% & 71.6\% \\
  4 & Moirai 2.0 (Salesforce) & 12 & 64.0\% & 64.0\% & 64.0\% & 69.6\% & 67.1\% & 69.6\% & 68.8\% \\
  5 & Chronos-2 (Amazon) & 120 & 64.6\% & 64.6\% & 64.6\% & 66.8\% & 63.0\% & 66.8\% & 68.5\% \\
  6 & TimesFM 500M (Google) & 500 & 58.7\% & 58.7\% & 58.7\% & 68.5\% & 65.9\% & 68.5\% & 66.7\% \\
  7 & PatchTST-FM (IBM) & 260 & 62.8\% & 62.8\% & 62.8\% & 63.8\% & 62.1\% & 63.9\% & 65.0\% \\
  8 & Chronos-Bolt-Mini (Amazon) & 21 & 56.1\% & 56.1\% & 56.1\% & 66.5\% & 65.2\% & 66.6\% & 66.3\% \\
  9 & TimesFM 200M (Google) & 200 & --- & --- & --- & 63.6\% & 63.0\% & 63.6\% & 62.1\% \\
  10 & Chronos-Bolt-Tiny (Amazon) & 9 & 56.7\% & 56.7\% & 56.7\% & 66.0\% & 65.0\% & 66.1\% & 65.7\% \\
  11 & Chronos-2-Small (Amazon) & 40 & 60.3\% & 60.3\% & 60.3\% & 63.4\% & 61.1\% & 63.5\% & 65.0\% \\
  12 & Chronos-Bolt-Base (Amazon) & 205 & 55.5\% & 55.5\% & 55.5\% & 64.5\% & 64.7\% & 64.6\% & 63.9\% \\
  13 & Chronos-Bolt-Small (Amazon) & 48 & 52.7\% & 52.7\% & 52.7\% & 63.7\% & 62.4\% & 63.8\% & 63.3\% \\
  14 & Moirai-Base (Salesforce) & 91 & 60.1\% & 60.1\% & 60.1\% & 58.5\% & 56.2\% & 58.5\% & 60.4\% \\
  15 & TabPFN-TS (Prior Labs) & 11 & 50.3\% & 50.3\% & 50.3\% & 62.6\% & 59.9\% & 62.6\% & 61.7\% \\
  16 & Chronos-Large (Amazon) & 710 & 45.2\% & 45.2\% & 45.2\% & 62.6\% & 63.7\% & 62.7\% & 61.6\% \\
  17 & Moirai-Large (Salesforce) & 311 & 56.6\% & 56.6\% & 56.6\% & 57.7\% & 55.1\% & 57.8\% & 58.7\% \\
  18 & Chronos-Mini (Amazon) & 20 & 45.4\% & 45.4\% & 45.4\% & 62.6\% & 62.0\% & 62.5\% & 60.7\% \\
  19 & Granite FlowState (IBM) & 9.1 & 48.9\% & 48.9\% & 48.9\% & 58.2\% & 59.0\% & 58.2\% & 58.1\% \\
  20 & Moirai-Small (Salesforce) & 14 & 54.4\% & 54.4\% & 54.4\% & 56.2\% & 52.9\% & 56.2\% & 57.5\% \\
  21 & Chronos-Tiny (Amazon) & 8 & 44.4\% & 44.4\% & 44.4\% & 59.1\% & 61.1\% & 59.0\% & 60.0\% \\
  22 & N-HiTS (Nixtla) & — & --- & --- & --- & 56.2\% & 55.3\% & 55.4\% & 51.9\% \\
  23 & Kairos-50M (Foundation Model Research) & 50 & 45.8\% & 45.8\% & 45.8\% & 57.9\% & 56.4\% & 57.9\% & 56.8\% \\
  24 & Chronos-Small (Amazon) & 46 & 41.6\% & 41.6\% & 41.6\% & 58.0\% & 57.7\% & 58.0\% & 59.2\% \\
  25 & N-BEATS (Nixtla) & — & --- & --- & --- & 53.1\% & 53.9\% & 53.1\% & 52.3\% \\
  26 & TOTO (Datadog) & 151 & 63.2\% & 63.2\% & 63.2\% & 50.3\% & 51.3\% & 50.4\% & 40.3\% \\
  27 & Kairos-10M (Foundation Model Research) & 10 & 42.0\% & 42.0\% & 42.0\% & 55.2\% & 55.7\% & 55.3\% & 55.6\% \\
  28 & Chronos-Base (Amazon) & 200 & 38.5\% & 38.5\% & 38.5\% & 56.1\% & 58.0\% & 56.1\% & 55.1\% \\
  29 & LSTM (Nixtla) & — & --- & --- & --- & 50.4\% & 49.7\% & 50.2\% & 47.2\% \\
  30 & Moirai-MoE (Salesforce) & 86 & 49.5\% & 49.5\% & 49.5\% & 49.6\% & 48.9\% & 49.7\% & 48.4\% \\
  31 & Kairos-23M (Foundation Model Research) & 23 & 40.1\% & 40.1\% & 40.1\% & 51.2\% & 49.7\% & 51.2\% & 53.0\% \\
  32 & PatchTSMixer (IBM) & — & --- & --- & --- & 43.4\% & 43.9\% & 43.6\% & 44.7\% \\
  33 & iTransformer (THUML) & — & --- & --- & --- & 44.2\% & 43.1\% & 44.0\% & 42.7\% \\
  34 & SVR (scikit-learn) & — & --- & --- & --- & 43.8\% & 41.9\% & 43.8\% & 44.3\% \\
  35 & Random Forest (scikit-learn) & — & --- & --- & --- & 43.3\% & 42.6\% & 43.2\% & 43.5\% \\
  36 & ARIMA (Statsmodels) & — & --- & --- & --- & 43.2\% & 41.8\% & 42.1\% & 41.0\% \\
  37 & Exponential Smoothing (Statsmodels) & — & --- & --- & --- & 42.1\% & 40.1\% & 42.1\% & 40.5\% \\
  38 & Croston (Statsmodels) & — & --- & --- & --- & 39.4\% & 38.9\% & 40.3\% & 38.0\% \\
  39 & TTM R2 (IBM) & 1.7 & --- & --- & --- & 36.2\% & 41.4\% & 36.1\% & 39.3\% \\
  40 & TTM R1 (IBM) & 1 & --- & --- & --- & 36.0\% & 41.2\% & 35.9\% & 39.2\% \\
  41 & Time-MoE-200M (Shanghai AI Lab) & 200 & --- & --- & --- & 35.5\% & 41.0\% & 35.5\% & 38.8\% \\
  42 & Time-MoE-50M (Shanghai AI Lab) & 50 & --- & --- & --- & 35.5\% & 41.0\% & 35.5\% & 38.8\% \\
  43 & TTM R2.1 (IBM) & 1.7 & --- & --- & --- & 35.4\% & 40.3\% & 35.2\% & 38.8\% \\
  44 & MOMENT-Large (CMU) & 385 & --- & --- & --- & 35.5\% & 39.4\% & 35.3\% & 37.9\% \\
  45 & XGBoost (DMLC) & — & --- & --- & --- & 35.5\% & 36.2\% & 36.5\% & 37.0\% \\
  46 & Prophet (Meta) & — & --- & --- & --- & 35.4\% & 36.8\% & 35.4\% & 37.3\% \\
  47 & MOMENT-Base (CMU) & 125 & --- & --- & --- & 33.6\% & 36.6\% & 33.3\% & 35.5\% \\
  48 & MOMENT-Small (CMU) & 40 & --- & --- & --- & 31.3\% & 34.8\% & 31.2\% & 32.9\% \\
  49 & Seasonal Naive (Sktime) & — & --- & --- & --- & 29.4\% & 29.8\% & 29.7\% & 27.6\% \\
  50 & Lag-Llama (Nixtla) & 10 & 18.6\% & 18.6\% & 18.6\% & 26.0\% & 29.3\% & 26.1\% & 25.1\% \\
  51 & PatchTST-Granite (IBM) & 40 & 15.4\% & 15.4\% & 15.4\% & 25.7\% & 27.6\% & 25.6\% & 30.3\% \\
  52 & Theta (Sktime) & — & --- & --- & --- & 21.7\% & 23.6\% & 21.9\% & 23.2\% \\
  53 & Sundial (THUML) & 128 & 3.3\% & 3.3\% & 3.3\% & 3.1\% & 3.2\% & 3.1\% & 3.0\% \\
\bottomrule
\end{tabular}%
}
\end{table}

\begin{table}[!htpb]
\centering
\caption{\label{tab:winrate_multivariate}
Multivariate tasks -- Average win rates by metric for deterministic \& probabilistic models}
\resizebox{1\textwidth}{!}{%
\begin{tabular}{r|l|c|ccc|cccc}
\toprule
  \multirow{2}{*}{Rank} & \multirow{2}{*}{Model} & \multirow{2}{*}{Params (M)} & \multicolumn{3}{c|}{Stochastic} & \multicolumn{4}{c}{Deterministic} \\
  & & & CRPS & QS & WIS & MAE & MAPE & MASE & RMSE \\
\midrule
  1 & TiRex (NX-AI) & 300 & 78.1\% & 78.1\% & 78.1\% & \cellcolor{green!15}\textbf{86.2\%} & \cellcolor{green!15}\textbf{80.6\%} & \cellcolor{green!15}\textbf{86.2\%} & \cellcolor{green!15}\textbf{85.9\%} \\
  2 & TiRex 1.1 (NX-AI) & 300 & 78.3\% & 78.3\% & 78.3\% & 85.3\% & 78.9\% & 85.3\% & 84.8\% \\
  3 & TimesFM 2.5 (Google) & 200 & 73.3\% & 73.3\% & 73.3\% & 83.6\% & 76.7\% & 83.5\% & 82.5\% \\
  4 & Chronos-2 (Amazon) & 120 & 77.8\% & 77.8\% & 77.8\% & 81.3\% & 64.5\% & 81.1\% & 83.6\% \\
  5 & Chronos-2-Small (Amazon) & 40 & 74.8\% & 74.8\% & 74.8\% & 75.9\% & 66.7\% & 75.7\% & 81.0\% \\
  6 & TimesFM 500M (Google) & 500 & 66.0\% & 66.0\% & 66.0\% & 79.6\% & 67.1\% & 79.6\% & 77.8\% \\
  7 & PatchTST-FM (IBM) & 260 & 73.3\% & 73.3\% & 73.3\% & 76.0\% & 60.2\% & 75.9\% & 76.8\% \\
  8 & TabPFN-TS (Prior Labs) & 11 & 55.9\% & 55.9\% & 55.9\% & 72.6\% & 67.1\% & 72.3\% & 72.2\% \\
  9 & Chronos-Bolt-Mini (Amazon) & 21 & 53.6\% & 53.6\% & 53.6\% & 74.5\% & 65.8\% & 74.3\% & 72.1\% \\
  10 & TimesFM 200M (Google) & 200 & --- & --- & --- & 68.9\% & 60.4\% & 69.0\% & 66.1\% \\
  11 & Chronos-Bolt-Small (Amazon) & 48 & 55.7\% & 55.7\% & 55.7\% & 73.6\% & 57.4\% & 73.5\% & 75.0\% \\
  12 & Chronos-Bolt-Base (Amazon) & 205 & 53.2\% & 53.2\% & 53.2\% & 72.2\% & 64.8\% & 72.1\% & 72.5\% \\
  13 & Chronos-Bolt-Tiny (Amazon) & 9 & 47.8\% & 47.8\% & 47.8\% & 70.0\% & 59.8\% & 69.9\% & 71.7\% \\
  14 & Moirai 2.0 (Salesforce) & 12 & 51.2\% & 51.2\% & 51.2\% & 65.3\% & 56.1\% & 65.0\% & 66.0\% \\
  15 & Kairos-50M (Foundation Model Research) & 50 & 48.1\% & 48.1\% & 48.1\% & 60.5\% & 67.5\% & 60.4\% & 57.8\% \\
  16 & Kairos-23M (Foundation Model Research) & 23 & 43.3\% & 43.3\% & 43.3\% & 62.6\% & 58.2\% & 62.6\% & 57.8\% \\
  17 & PatchTSMixer (IBM) & — & --- & --- & --- & 52.9\% & 50.2\% & 52.4\% & 56.6\% \\
  18 & Granite FlowState (IBM) & 9.1 & 41.2\% & 41.2\% & 41.2\% & 56.6\% & 60.4\% & 56.2\% & 58.3\% \\
  19 & N-HiTS (Nixtla) & — & --- & --- & --- & 55.4\% & 36.9\% & 54.6\% & 55.3\% \\
  20 & Kairos-10M (Foundation Model Research) & 10 & 35.1\% & 35.1\% & 35.1\% & 54.0\% & 61.5\% & 54.0\% & 56.5\% \\
  21 & Chronos-Mini (Amazon) & 20 & 35.3\% & 35.3\% & 35.3\% & 58.1\% & 46.8\% & 57.8\% & 56.2\% \\
  22 & Chronos-Base (Amazon) & 200 & 35.7\% & 35.7\% & 35.7\% & 58.9\% & 46.8\% & 58.8\% & 52.1\% \\
  23 & N-BEATS (Nixtla) & — & --- & --- & --- & 51.9\% & 32.5\% & 52.8\% & 53.9\% \\
  24 & Chronos-Small (Amazon) & 46 & 34.1\% & 34.1\% & 34.1\% & 56.2\% & 50.4\% & 56.1\% & 50.4\% \\
  25 & Chronos-Tiny (Amazon) & 8 & 32.9\% & 32.9\% & 32.9\% & 54.1\% & 46.2\% & 54.2\% & 52.6\% \\
  26 & Chronos-Large (Amazon) & 710 & 32.9\% & 32.9\% & 32.9\% & 52.9\% & 44.8\% & 52.7\% & 48.9\% \\
  27 & TOTO (Datadog) & 151 & \cellcolor{green!15}\textbf{78.7\%} & \cellcolor{green!15}\textbf{78.7\%} & \cellcolor{green!15}\textbf{78.7\%} & 34.4\% & 34.8\% & 34.5\% & 19.5\% \\
  28 & iTransformer (THUML) & — & --- & --- & --- & 45.5\% & 38.4\% & 45.2\% & 45.1\% \\
  29 & Moirai-Large (Salesforce) & 311 & 56.4\% & 56.4\% & 56.4\% & 39.8\% & 33.6\% & 39.7\% & 38.9\% \\
  30 & SVR (scikit-learn) & — & --- & --- & --- & 42.1\% & 44.9\% & 42.1\% & 44.0\% \\
  31 & Prophet (Meta) & — & --- & --- & --- & 38.9\% & 47.2\% & 38.5\% & 44.6\% \\
  32 & TTM R2 (IBM) & 1.7 & --- & --- & --- & 36.0\% & 56.6\% & 35.8\% & 37.8\% \\
  33 & TTM R2.1 (IBM) & 1.7 & --- & --- & --- & 35.7\% & 55.4\% & 35.5\% & 38.0\% \\
  34 & Croston (Statsmodels) & — & --- & --- & --- & 39.1\% & 45.7\% & 39.7\% & 39.2\% \\
  35 & TTM R1 (IBM) & 1 & --- & --- & --- & 35.3\% & 55.7\% & 35.1\% & 37.2\% \\
  36 & Moirai-Base (Salesforce) & 91 & 49.1\% & 49.1\% & 49.1\% & 37.4\% & 33.0\% & 37.2\% & 34.2\% \\
  37 & MOMENT-Large (CMU) & 385 & --- & --- & --- & 35.5\% & 49.1\% & 35.2\% & 36.6\% \\
  38 & ARIMA (Statsmodels) & — & --- & --- & --- & 33.2\% & 48.0\% & 34.4\% & 37.8\% \\
  39 & Random Forest (scikit-learn) & — & --- & --- & --- & 35.6\% & 41.1\% & 35.4\% & 38.5\% \\
  40 & Time-MoE-200M (Shanghai AI Lab) & 200 & --- & --- & --- & 29.2\% & 52.5\% & 29.1\% & 33.7\% \\
  41 & Time-MoE-50M (Shanghai AI Lab) & 50 & --- & --- & --- & 29.2\% & 52.5\% & 29.1\% & 33.7\% \\
  42 & MOMENT-Base (CMU) & 125 & --- & --- & --- & 31.8\% & 45.9\% & 31.6\% & 32.3\% \\
  43 & LSTM (Nixtla) & — & --- & --- & --- & 32.8\% & 38.7\% & 35.8\% & 30.7\% \\
  44 & MOMENT-Small (CMU) & 40 & --- & --- & --- & 30.4\% & 42.4\% & 30.2\% & 30.5\% \\
  45 & Moirai-Small (Salesforce) & 14 & 31.5\% & 31.5\% & 31.5\% & 33.2\% & 30.6\% & 33.0\% & 32.4\% \\
  46 & Exponential Smoothing (Statsmodels) & — & --- & --- & --- & 27.2\% & 35.4\% & 29.7\% & 30.0\% \\
  47 & Seasonal Naive (Sktime) & — & --- & --- & --- & 26.7\% & 39.5\% & 25.4\% & 24.1\% \\
  48 & PatchTST-Granite (IBM) & 40 & 6.6\% & 6.7\% & 6.7\% & 18.0\% & 22.6\% & 17.8\% & 23.4\% \\
  49 & Theta (Sktime) & — & --- & --- & --- & 13.5\% & 24.2\% & 13.5\% & 12.7\% \\
  50 & Sundial (THUML) & 128 & 0.0\% & 0.0\% & 0.0\% & 0.2\% & 3.9\% & 0.2\% & 0.6\% \\
\bottomrule
\end{tabular}%
}
\end{table}

\begin{table}[!htpb]
\centering
\caption{\label{tab:winrate_covariate}
Covariate tasks -- Average win rates by metric for deterministic \& probabilistic models}
\resizebox{1\textwidth}{!}{%
\begin{tabular}{r|l|c|ccc|cccc}
\toprule
  \multirow{2}{*}{Rank} & \multirow{2}{*}{Model} & \multirow{2}{*}{Params (M)} & \multicolumn{3}{c|}{Stochastic} & \multicolumn{4}{c}{Deterministic} \\
  & & & CRPS & QS & WIS & MAE & MAPE & MASE & RMSE \\
\midrule
  1 & Chronos-2 (Amazon) & 120 & \cellcolor{green!15}\textbf{83.0\%} & \cellcolor{green!15}\textbf{83.0\%} & \cellcolor{green!15}\textbf{83.0\%} & 79.1\% & 72.1\% & 79.3\% & \cellcolor{green!15}\textbf{80.7\%} \\
  2 & TabPFN-TS (Prior Labs) & 11 & 77.0\% & 77.0\% & 77.0\% & \cellcolor{green!15}\textbf{82.0\%} & \cellcolor{green!15}\textbf{80.4\%} & \cellcolor{green!15}\textbf{82.2\%} & 72.3\% \\
  3 & Chronos-2-Small (Amazon) & 40 & 80.3\% & 80.3\% & 80.3\% & 74.1\% & 62.3\% & 74.2\% & 79.7\% \\
  4 & TimesFM 200M (Google) & 200 & --- & --- & --- & 72.2\% & 71.0\% & 72.1\% & 65.4\% \\
  5 & Chronos-Bolt-Base (Amazon) & 205 & 61.1\% & 61.1\% & 61.1\% & 67.2\% & 69.3\% & 67.2\% & 67.9\% \\
  6 & Chronos-Bolt-Tiny (Amazon) & 9 & 54.8\% & 54.8\% & 54.8\% & 69.6\% & 71.1\% & 69.7\% & 70.0\% \\
  7 & Prophet (Meta) & — & --- & --- & --- & 64.3\% & 62.5\% & 64.6\% & 69.9\% \\
  8 & Chronos-Bolt-Small (Amazon) & 48 & 57.7\% & 57.7\% & 57.7\% & 68.0\% & 67.9\% & 68.0\% & 69.4\% \\
  9 & Chronos-Bolt-Mini (Amazon) & 21 & 58.0\% & 58.0\% & 58.0\% & 67.1\% & 68.3\% & 67.2\% & 68.8\% \\
  10 & TiRex (NX-AI) & 300 & 62.4\% & 62.4\% & 62.4\% & 67.3\% & 63.6\% & 67.5\% & 60.8\% \\
  11 & TiRex 1.1 (NX-AI) & 300 & 60.5\% & 60.5\% & 60.5\% & 67.5\% & 63.2\% & 67.8\% & 61.5\% \\
  12 & TimesFM 2.5 (Google) & 200 & 55.6\% & 55.6\% & 55.6\% & 65.3\% & 61.7\% & 65.4\% & 62.4\% \\
  13 & PatchTST-FM (IBM) & 260 & 59.1\% & 59.1\% & 59.1\% & 58.4\% & 61.5\% & 58.4\% & 65.9\% \\
  14 & Moirai 2.0 (Salesforce) & 12 & 54.5\% & 54.5\% & 54.5\% & 65.7\% & 62.0\% & 65.8\% & 57.5\% \\
  15 & Kairos-50M (Foundation Model Research) & 50 & 55.0\% & 55.0\% & 55.0\% & 62.3\% & 61.1\% & 62.4\% & 59.0\% \\
  16 & ARIMA (Statsmodels) & — & --- & --- & --- & 58.9\% & 56.1\% & 57.0\% & 62.3\% \\
  17 & Kairos-10M (Foundation Model Research) & 10 & 49.0\% & 49.0\% & 49.0\% & 60.6\% & 58.4\% & 60.6\% & 57.0\% \\
  18 & TimesFM 500M (Google) & 500 & 52.1\% & 52.1\% & 52.1\% & 58.2\% & 57.7\% & 58.3\% & 55.0\% \\
  19 & Moirai-Small (Salesforce) & 14 & 54.6\% & 54.6\% & 54.6\% & 51.4\% & 53.4\% & 51.5\% & 55.9\% \\
  20 & Kairos-23M (Foundation Model Research) & 23 & 43.4\% & 43.4\% & 43.4\% & 57.5\% & 55.3\% & 57.6\% & 54.9\% \\
  21 & Chronos-Mini (Amazon) & 20 & 34.1\% & 34.1\% & 34.1\% & 61.1\% & 62.5\% & 61.0\% & 53.6\% \\
  22 & PatchTSMixer (IBM) & — & --- & --- & --- & 49.2\% & 52.7\% & 48.1\% & 56.3\% \\
  23 & Granite FlowState (IBM) & 9.1 & 42.3\% & 42.3\% & 42.3\% & 56.2\% & 54.1\% & 56.3\% & 52.9\% \\
  24 & TOTO (Datadog) & 151 & 57.5\% & 57.5\% & 57.5\% & 45.6\% & 47.1\% & 45.8\% & 41.5\% \\
  25 & Chronos-Tiny (Amazon) & 8 & 31.2\% & 31.2\% & 31.2\% & 57.7\% & 55.8\% & 57.7\% & 49.7\% \\
  26 & Chronos-Base (Amazon) & 200 & 33.2\% & 33.2\% & 33.2\% & 53.7\% & 54.2\% & 53.7\% & 55.4\% \\
  27 & Exponential Smoothing (Statsmodels) & — & --- & --- & --- & 45.2\% & 47.7\% & 45.6\% & 53.1\% \\
  28 & Chronos-Small (Amazon) & 46 & 26.9\% & 26.9\% & 26.9\% & 55.8\% & 55.3\% & 56.0\% & 51.6\% \\
  29 & Moirai-Base (Salesforce) & 91 & 48.7\% & 48.7\% & 48.7\% & 44.6\% & 44.3\% & 45.0\% & 48.0\% \\
  30 & LSTM (Nixtla) & — & --- & --- & --- & 48.8\% & 49.0\% & 47.3\% & 33.9\% \\
  31 & Croston (Statsmodels) & — & --- & --- & --- & 44.0\% & 42.7\% & 45.6\% & 46.3\% \\
  32 & TTM R2 (IBM) & 1.7 & --- & --- & --- & 39.5\% & 47.2\% & 39.5\% & 45.4\% \\
  33 & TTM R1 (IBM) & 1 & --- & --- & --- & 38.6\% & 47.1\% & 38.7\% & 45.2\% \\
  34 & MOMENT-Large (CMU) & 385 & --- & --- & --- & 38.3\% & 44.9\% & 38.1\% & 44.3\% \\
  35 & iTransformer (THUML) & — & --- & --- & --- & 38.4\% & 36.6\% & 38.5\% & 41.6\% \\
  36 & Random Forest (scikit-learn) & — & --- & --- & --- & 34.1\% & 39.3\% & 34.2\% & 39.7\% \\
  37 & Time-MoE-200M (Shanghai AI Lab) & 200 & --- & --- & --- & 33.2\% & 40.3\% & 33.3\% & 38.6\% \\
  38 & Time-MoE-50M (Shanghai AI Lab) & 50 & --- & --- & --- & 33.2\% & 40.3\% & 33.3\% & 38.6\% \\
  39 & SVR (scikit-learn) & — & --- & --- & --- & 36.3\% & 35.6\% & 36.4\% & 36.6\% \\
  40 & MOMENT-Small (CMU) & 40 & --- & --- & --- & 35.4\% & 35.4\% & 35.3\% & 38.2\% \\
  41 & N-BEATS (Nixtla) & — & --- & --- & --- & 37.1\% & 35.2\% & 38.1\% & 33.2\% \\
  42 & MOMENT-Base (CMU) & 125 & --- & --- & --- & 33.7\% & 37.1\% & 33.8\% & 37.8\% \\
  43 & Seasonal Naive (Sktime) & — & --- & --- & --- & 33.3\% & 29.7\% & 31.9\% & 27.0\% \\
  44 & PatchTST-Granite (IBM) & 40 & 8.1\% & 8.1\% & 8.1\% & 27.3\% & 26.4\% & 27.2\% & 32.7\% \\
  45 & N-HiTS (Nixtla) & — & --- & --- & --- & 7.5\% & 6.5\% & 7.1\% & 6.9\% \\
  46 & Theta (Sktime) & — & --- & --- & --- & 3.6\% & 1.9\% & 3.6\% & 3.8\% \\
  47 & Sundial (THUML) & 128 & 0.0\% & 0.0\% & 0.0\% & 1.9\% & 1.9\% & 1.9\% & 1.9\% \\
\bottomrule
\end{tabular}%
}
\end{table}

\subsection{Skill Score Results}

In Tables~\ref{tab:skill_scores_univariate}, \ref{tab:skill_scores_multivariate}, and \ref{tab:skill_scores_covariate} (Appendix~\ref{sec_app:more_results}), we report the per-metric averaged skill scores, across univariate, multivariate, and covariate forecasting tasks, respectively. 
All skill scores compare model performance to a baseline model (Seasonal Naive). Positive skill scores indicate better performance than the baseline, while negative scores indicate worse performance.

\paragraph{Univariate (see Table \ref{tab:skill_scores_univariate}).}
The strongest models (TiRex, TiRex~1.1, TimesFM~2.5, Moirai~2.0, Chronos-2 and smaller Chronos variants) achieve roughly $0.4$--$0.66$ skill on MAE, MAPE, and MASE---substantial geometric-mean improvement versus naive on those errors. TiRex leads MAE, MAPE, and MASE, while Chronos-2 leads RMSE, illustrating metric-dependent rankings. Many classical and lighter models remain above naive but with smaller skill (often about $0.1$--$0.4$). Several foundation-model families (TTM, Time-MoE, MOMENT) and some baselines (e.g.\ XGBoost, Lag-Llama, PatchTST-Granite, Theta) fall below zero on most columns. Sundial exhibits extremely large negative skill, which is more suggestive of severe mismatch or unstable ratio behavior under clipping than of a mild underperformance relative to naive. TOTO illustrates heterogeneous metric behavior: decent skill on MAE, MAPE, and MASE but comparatively weak RMSE skill.

\paragraph{Multivariate (see Table \ref{tab:skill_scores_multivariate}).}
TimesFM~2.5 attains the best skill in all four columns on this slice, with values around $0.64$--$0.68$, slightly more consistent than in the univariate table. TiRex remains very competitive but does not win any column. Several models show mixed signs across metrics (e.g.\ Moirai~2.0 is positive on MAE, MASE, and RMSE but negative on MAPE; N-BEATS and TOTO show strongly negative MAPE while other metrics are mixed or positive), so MAPE-based conclusions should be drawn carefully. Moirai-Base, Moirai-Large, PatchTST-Granite, and Theta are deeply negative on multiple metrics; Sundial is again a large-magnitude outlier.

\paragraph{Covariate (see Table \ref{tab:skill_scores_covariate}).}
TabPFN-TS achieves the highest skill in every column (about $0.74$--$0.82$), ahead of Chronos-2 (about $0.63$--$0.69$), indicating particular strength when exogenous covariates are part of the task definition. TimesFM~200M also scores highly. Classical models such as ARIMA and Prophet occupy a strong mid-to-upper tier. As in the other slices, Time-MoE variants are clearly below naive on average for several metrics, and N-HiTS, Theta, and Sundial show extreme negative skill values that warrant separate discussion of failure modes rather than ranking alone.

\section{Future Directions and Conclusion}
\label{sec:conclusion}

In line with existing forecasting evaluation frameworks, in \tempusbench, we only consider benchmarks based on standard forecasting task features, such as target variate type, context length, and forecast length. Benchmarks could be defined more comprehensively in terms of hyperparameters that correspond to standard forecasting task features and others, such as task domains. Indeed, a more comprehensive set of benchmarks could test the per-domain (e.g., economics) performance of forecasting models across different target variate types, context lengths, and forecast lengths. In future work, we plan to develop a more flexible taxonomy for describing forecasting tasks.

Finally, we expect the datasets used to define our benchmarks will eventually be included in the pretaining corpora of TSFMs, as has been the case with many NLP benchmarks. To this end, in future work, we plan to develop dynamic benchmarks where test data is continuously refreshed. We have in mind two types of dynamic benchmarks: those based on synthetic data, where new datasets are periodically generated (e.g., seasonality tests), and those based on real-world data that update naturally over time (e.g., monthly inflation reports or daily weather observations).

\newpage
\bibliographystyle{unsrtnat}  
\bibliography{references}

\newpage
\appendix
\section{Additional Mathematical Background}
\label{sec_app:prelims}
\subsection{Mathematical notation}

We adopt the following conventions to indicate the nature of the mathematical object at hand: We use calligraphic uppercase letters to denote sets (e.g., $\calX$), bold uppercase letters to denote matrices (e.g., $\X$), bold lowercase letters to denote vectors (e.g., $\p$), lowercase letters to denote scalar quantities (e.g., $x$), and uppercase letters to denote random variables (e.g., $X$). 

We denote the $i$th row vector of a matrix (e.g., $\X$) by the corresponding bold lowercase letter with subscript $i$ (e.g., $\x_i)$. 
Similarly, we denote the $j$th entry of a vector (e.g., $\p$ or $\x_i$) by the corresponding lowercase letter with subscript $j$ (e.g., $p_j$ or $x_{ij}$).
We denote functions by a letter determined by the value of the function, e.g., $f$ if the mapping is scalar valued, $\f$ if the mapping is vector valued, and $\calF$ if the mapping is set valued.

We denote the set $\left\{1, \hdots, n\right\}$ by $[n]$, the set $\left\{n, n+1, \hdots, m\right\}$ by $[n:m]$, the set of natural numbers by $\N$, and the set of real numbers by $\R$. 
We denote the positive and strictly positive elements of a set using a $+$ or $++$ subscript, respectively, e.g., $\R_+$ and $\R_{++}$.
For any $n \in \N$, we denote the  $n$-dimensional vector of zeros and ones by $\zeros[n]$ and $\ones[n]$, respectively.

\subsection{Mathematical Definitions}

We let $\simplex[n] = \{\x \in \R_+^n \mid \sum_{i = 1}^n x_i = 1 \}$ denote the unit simplex in $\R^n$, and $\simplex(A)$ denote the set of all probability measures over a given set $A$.
We also define the support of a probability density function $f \in \simplex(\calX)$ as $\supp(f) \doteq \left\{ \x \in \calX \mid f(\x) > 0 \right\}$.
Finally, we denote the orthogonal projection operator onto a set $C$ by $\project[C]$, i.e., $\project[C](\x) \doteq \argmin_{\y \in C} \left\|\x - \y \right\|^2$.

\subsection{Evaluation Metrics}
An evaluation metric $\metric: \targetset^{\forecastlen} \times \targetset^{\forecastlen} \to \R_+$ is a positive-, scalar-valued function s.t. for any forecast $\forecastts \in \targetset^{\forecastlen}$ and realized future target values $\targetts[][][][*] \in \targetset^{\forecastlen}$, $\metric(\forecastts, \targetts[][][][*]) \geq 0$ denotes the distance between the forecast and the realized values. 
We consider the following evaluation metrics \samy{at present}{}. 

The \mydef{mean absolute error (MAE)} is defined as $$\metric[{\mathrm{MAE}}](\forecastts, \targetts[][][][*]) \doteq \frac{1}{\numtargets \forecastlen} \sum_{\target \in \targets} \sum_{\timestep = 1}^{\forecastlen} |\forecastts[\target][\timestep] - \targetts[\target][\timestep][][*] |$$

The \mydef{root mean squared error (RMSE)} is defined as $$\metric[{\mathrm{RMSE}}](\forecastts, \targetts[][][][*]) \doteq \sqrt{ \frac{1}{\numtargets \forecastlen} \sum_{\target \in \targets} \sum_{\timestep = 1}^{\forecastlen} (\forecastts[\target][\timestep] - \targetts[\target][\timestep][][*])^2 }$$

The \mydef{mean absolute scale error (MASE)} is defined as $$\metric[{\mathrm{MASE}}](\forecastts, \targetts[][][][*]) \doteq \frac{1}{\numtargets \forecastlen} \sum_{\target \in \targets} \sum_{\timestep = 1}^{\forecastlen} \frac{|\forecastts[\target][\timestep] - \targetts[\target][\timestep][][*] |}{\frac{1}{\forecastlen-1}\sum_{\timestep = 1}^{\contextlen-1} |\targetts[\target][\timestep + 1][][] - \targetts[\target][\timestep][][]|}$$

We note MAE is scale-dependent but less sensitive to outliers than MSE, which disproportionately penalizes large forecast errors.
MASE normalizes errors w.r.t.\@ the forecasts of a naive forecast method (i.e., setting the next time step's forecast to be the current time step's realized value), making it scale-free and comparable across datasets and domains.

The \mydef{mean absolute percentage error (MAPE)} is defined as $$\metric[{\mathrm{MAPE}}](\forecastts, \targetts[][][][*]) \doteq \frac{100}{\numtargets \forecastlen} \sum_{\target \in \targets} \sum_{\timestep = 1}^{\forecastlen} \frac{|\forecastts[\target][\timestep] - \targetts[\target][\timestep][][*] |}{|\targetts[\target][\timestep][][*]|}$$

\newpage
\section{Skill Score Results}
\label{sec_app:more_results}

\begingroup
\small
\setlength{\LTpre}{6pt}
\setlength{\LTpost}{6pt}
\begin{longtable}{r|l|cccc}
\caption{Per-metric averaged skill scores across univariate tasks}
\label{tab:skill_scores_univariate}\\
\toprule
  Rank & Model & MAE & MAPE & MASE & RMSE \\
\midrule
\endfirsthead
\multicolumn{6}{c}{\tablename~\thetable{} --- \textit{continued from previous page}} \\
\toprule
  Rank & Model & MAE & MAPE & MASE & RMSE \\
\midrule
\endhead
\midrule
\multicolumn{6}{r}{\textit{Continued on next page}} \\
\endfoot
\bottomrule
\endlastfoot
  1 & TiRex (NX-AI) & \cellcolor{green!15}\textbf{0.639} & \cellcolor{green!15}\textbf{0.660} & \cellcolor{green!15}\textbf{0.644} & 0.596 \\
  2 & TiRex 1.1 (NX-AI) & 0.625 & 0.635 & 0.631 & 0.579 \\
  3 & TimesFM 2.5 (Google) & 0.622 & 0.640 & 0.628 & 0.584 \\
  4 & Moirai 2.0 (Salesforce) & 0.621 & 0.639 & 0.627 & 0.578 \\
  5 & Chronos-2 (Amazon) & 0.637 & 0.644 & 0.642 & \cellcolor{green!15}\textbf{0.608} \\
  6 & TimesFM 500M (Google) & 0.525 & 0.563 & 0.532 & 0.482 \\
  7 & PatchTST-FM (IBM) & 0.506 & 0.518 & 0.513 & 0.475 \\
  8 & Chronos-Bolt-Mini (Amazon) & 0.551 & 0.581 & 0.558 & 0.517 \\
  9 & TimesFM 200M (Google) & 0.477 & 0.517 & 0.485 & 0.418 \\
  10 & Chronos-Bolt-Tiny (Amazon) & 0.544 & 0.570 & 0.551 & 0.510 \\
  11 & Chronos-2-Small (Amazon) & 0.623 & 0.631 & 0.629 & 0.592 \\
  12 & Chronos-Bolt-Base (Amazon) & 0.508 & 0.553 & 0.515 & 0.468 \\
  13 & Chronos-Bolt-Small (Amazon) & 0.526 & 0.557 & 0.534 & 0.492 \\
  14 & Moirai-Base (Salesforce) & 0.479 & 0.463 & 0.486 & 0.434 \\
  15 & TabPFN-TS (Prior Labs) & 0.516 & 0.536 & 0.523 & 0.463 \\
  16 & Chronos-Large (Amazon) & 0.437 & 0.510 & 0.445 & 0.389 \\
  17 & Moirai-Large (Salesforce) & 0.467 & 0.443 & 0.475 & 0.420 \\
  18 & Chronos-Mini (Amazon) & 0.438 & 0.517 & 0.446 & 0.391 \\
  19 & Granite FlowState (IBM) & 0.447 & 0.478 & 0.456 & 0.409 \\
  20 & Moirai-Small (Salesforce) & 0.430 & 0.408 & 0.438 & 0.399 \\
  21 & Chronos-Tiny (Amazon) & 0.417 & 0.500 & 0.426 & 0.371 \\
  22 & N-HiTS (Nixtla) & 0.432 & 0.433 & 0.415 & 0.354 \\
  23 & Kairos-50M (Foundation Model Research) & 0.459 & 0.498 & 0.468 & 0.422 \\
  24 & Chronos-Small (Amazon) & 0.378 & 0.423 & 0.387 & 0.341 \\
  25 & N-BEATS (Nixtla) & 0.378 & 0.421 & 0.387 & 0.332 \\
  26 & TOTO (Datadog) & 0.364 & 0.382 & 0.373 & 0.128 \\
  27 & Kairos-10M (Foundation Model Research) & 0.462 & 0.521 & 0.470 & 0.431 \\
  28 & Chronos-Base (Amazon) & 0.360 & 0.437 & 0.370 & 0.324 \\
  29 & LSTM (Nixtla) & 0.127 & 0.248 & 0.192 & 0.040 \\
  30 & Moirai-MoE (Salesforce) & 0.309 & 0.295 & 0.319 & 0.256 \\
  31 & Kairos-23M (Foundation Model Research) & 0.383 & 0.406 & 0.392 & 0.363 \\
  32 & PatchTSMixer (IBM) & 0.084 & 0.127 & 0.096 & 0.077 \\
  33 & iTransformer (THUML) & 0.226 & 0.215 & 0.237 & 0.171 \\
  34 & SVR (scikit-learn) & 0.172 & 0.217 & 0.185 & 0.146 \\
  35 & Random Forest (scikit-learn) & 0.124 & 0.172 & 0.137 & 0.123 \\
  36 & ARIMA (Statsmodels) & 0.172 & 0.217 & 0.155 & 0.138 \\
  37 & Exponential Smoothing (Statsmodels) & 0.093 & 0.083 & 0.099 & 0.049 \\
  38 & Croston (Statsmodels) & 0.049 & 0.106 & 0.039 & 0.015 \\
  39 & TTM R2 (IBM) & -0.114 & -0.018 & -0.097 & -0.079 \\
  40 & TTM R1 (IBM) & -0.117 & -0.018 & -0.100 & -0.082 \\
  41 & Time-MoE-200M (Shanghai AI Lab) & -0.145 & -0.023 & -0.127 & -0.108 \\
  42 & Time-MoE-50M (Shanghai AI Lab) & -0.145 & -0.023 & -0.127 & -0.108 \\
  43 & TTM R2.1 (IBM) & -0.119 & -0.019 & -0.102 & -0.084 \\
  44 & MOMENT-Large (CMU) & -0.118 & -0.030 & -0.102 & -0.084 \\
  45 & XGBoost (DMLC) & -0.106 & -0.001 & -0.056 & -0.064 \\
  46 & Prophet (Meta) & 0.034 & 0.086 & 0.049 & 0.064 \\
  47 & MOMENT-Base (CMU) & -0.127 & -0.059 & -0.110 & -0.095 \\
  48 & MOMENT-Small (CMU) & -0.138 & -0.072 & -0.121 & -0.108 \\
  49 & Seasonal Naive (Sktime) & 0.000 & 0.000 & 0.000 & 0.000 \\
  50 & Lag-Llama (Nixtla) & -0.650 & -0.426 & -0.626 & -0.620 \\
  51 & PatchTST-Granite (IBM) & -0.434 & -0.457 & -0.413 & -0.371 \\
  52 & Theta (Sktime) & -0.647 & -0.632 & -0.622 & -0.570 \\
  53 & Sundial (THUML) & -91.679 & -83.721 & -90.430 & -96.323 \\
\end{longtable}
\endgroup

\begingroup
\small
\setlength{\LTpre}{6pt}
\setlength{\LTpost}{6pt}
\begin{longtable}{r|l|cccc}
\caption{Per-metric averaged skill scores across multivariate tasks}
\label{tab:skill_scores_multivariate}\\
\toprule
  Rank & Model & MAE & MAPE & MASE & RMSE \\
\midrule
\endfirsthead
\multicolumn{6}{c}{\tablename~\thetable{} --- \textit{continued from previous page}} \\
\toprule
  Rank & Model & MAE & MAPE & MASE & RMSE \\
\midrule
\endhead
\midrule
\multicolumn{6}{r}{\textit{Continued on next page}} \\
\endfoot
\bottomrule
\endlastfoot
  1 & TiRex (NX-AI) & 0.629 & 0.462 & 0.639 & 0.617 \\
  2 & TiRex 1.1 (NX-AI) & 0.624 & 0.494 & 0.635 & 0.605 \\
  3 & TimesFM 2.5 (Google) & \cellcolor{green!15}\textbf{0.653} & \cellcolor{green!15}\textbf{0.681} & \cellcolor{green!15}\textbf{0.663} & \cellcolor{green!15}\textbf{0.644} \\
  4 & Chronos-2 (Amazon) & 0.524 & 0.154 & 0.537 & 0.523 \\
  5 & Chronos-2-Small (Amazon) & 0.573 & 0.417 & 0.585 & 0.576 \\
  6 & TimesFM 500M (Google) & 0.577 & 0.264 & 0.589 & 0.563 \\
  7 & PatchTST-FM (IBM) & 0.565 & 0.169 & 0.577 & 0.572 \\
  8 & TabPFN-TS (Prior Labs) & 0.532 & 0.530 & 0.545 & 0.523 \\
  9 & Chronos-Bolt-Mini (Amazon) & 0.479 & 0.194 & 0.494 & 0.471 \\
  10 & TimesFM 200M (Google) & 0.602 & 0.250 & 0.613 & 0.606 \\
  11 & Chronos-Bolt-Small (Amazon) & 0.484 & 0.137 & 0.498 & 0.482 \\
  12 & Chronos-Bolt-Base (Amazon) & 0.468 & 0.202 & 0.483 & 0.466 \\
  13 & Chronos-Bolt-Tiny (Amazon) & 0.474 & 0.152 & 0.489 & 0.473 \\
  14 & Moirai 2.0 (Salesforce) & 0.368 & -0.182 & 0.386 & 0.356 \\
  15 & Kairos-50M (Foundation Model Research) & 0.378 & 0.251 & 0.395 & 0.345 \\
  16 & Kairos-23M (Foundation Model Research) & 0.419 & 0.073 & 0.436 & 0.395 \\
  17 & PatchTSMixer (IBM) & 0.288 & -0.255 & 0.283 & 0.299 \\
  18 & Granite FlowState (IBM) & 0.437 & 0.088 & 0.453 & 0.430 \\
  19 & N-HiTS (Nixtla) & 0.360 & -0.344 & 0.379 & 0.385 \\
  20 & Kairos-10M (Foundation Model Research) & 0.357 & 0.091 & 0.375 & 0.350 \\
  21 & Chronos-Mini (Amazon) & 0.394 & -0.003 & 0.411 & 0.347 \\
  22 & Chronos-Base (Amazon) & 0.434 & -0.025 & 0.450 & 0.385 \\
  23 & N-BEATS (Nixtla) & 0.296 & -0.603 & 0.319 & 0.337 \\
  24 & Chronos-Small (Amazon) & 0.428 & 0.063 & 0.444 & 0.409 \\
  25 & Chronos-Tiny (Amazon) & 0.366 & 0.033 & 0.384 & 0.335 \\
  26 & Chronos-Large (Amazon) & 0.395 & -0.007 & 0.412 & 0.355 \\
  27 & TOTO (Datadog) & 0.260 & -0.876 & 0.281 & -0.345 \\
  28 & iTransformer (THUML) & 0.297 & -0.324 & 0.316 & 0.292 \\
  29 & Moirai-Large (Salesforce) & -0.381 & -2.120 & -0.343 & -1.358 \\
  30 & SVR (scikit-learn) & -0.047 & -0.143 & -0.018 & 0.052 \\
  31 & Prophet (Meta) & 0.313 & 0.400 & 0.332 & 0.375 \\
  32 & TTM R2 (IBM) & -0.140 & 0.203 & -0.108 & -0.096 \\
  33 & TTM R2.1 (IBM) & -0.139 & 0.195 & -0.107 & -0.096 \\
  34 & Croston (Statsmodels) & 0.092 & 0.196 & 0.124 & 0.152 \\
  35 & TTM R1 (IBM) & -0.140 & 0.201 & -0.108 & -0.096 \\
  36 & Moirai-Base (Salesforce) & -0.905 & -2.197 & -0.852 & -1.969 \\
  37 & MOMENT-Large (CMU) & -0.142 & -0.508 & -0.110 & -0.099 \\
  38 & ARIMA (Statsmodels) & 0.036 & 0.143 & 0.090 & 0.117 \\
  39 & Random Forest (scikit-learn) & -0.088 & -0.087 & -0.057 & -0.029 \\
  40 & Time-MoE-200M (Shanghai AI Lab) & -0.383 & 0.063 & -0.345 & -0.279 \\
  41 & Time-MoE-50M (Shanghai AI Lab) & -0.383 & 0.063 & -0.345 & -0.279 \\
  42 & MOMENT-Base (CMU) & -0.156 & -0.594 & -0.123 & -0.114 \\
  43 & LSTM (Nixtla) & -0.157 & -0.840 & -0.065 & -0.113 \\
  44 & MOMENT-Small (CMU) & -0.169 & -0.639 & -0.136 & -0.129 \\
  45 & Moirai-Small (Salesforce) & 0.091 & -2.152 & 0.117 & -0.156 \\
  46 & Exponential Smoothing (Statsmodels) & -0.058 & -0.099 & -0.046 & 0.001 \\
  47 & Seasonal Naive (Sktime) & 0.000 & 0.000 & 0.000 & 0.000 \\
  48 & PatchTST-Granite (IBM) & -0.483 & -2.102 & -0.442 & -0.293 \\
  49 & Theta (Sktime) & -1.640 & -1.427 & -1.566 & -1.166 \\
  50 & Sundial (THUML) & -93.691 & -92.964 & -93.798 & -94.665 \\
\end{longtable}
\endgroup

\begingroup
\small
\setlength{\LTpre}{6pt}
\setlength{\LTpost}{6pt}
\begin{longtable}{r|l|cccc}
\caption{Per-metric averaged skill scores across covariate tasks}
\label{tab:skill_scores_covariate}\\
\toprule
  Rank & Model & MAE & MAPE & MwASE & RMSE \\
\midrule
\endfirsthead
\multicolumn{6}{c}{\tablename~\thetable{} --- \textit{continued from previous page}} \\
\toprule
  Rank & Model & MAE & MAPE & MASE & RMSE \\
\midrule
\endhead
\midrule
\multicolumn{6}{r}{\textit{Continued on next page}} \\
\endfoot
\bottomrule
\endlastfoot
  1 & Chronos-2 (Amazon) & 0.630 & 0.688 & 0.638 & 0.628 \\
  2 & TabPFN-TS (Prior Labs) & \cellcolor{green!15}\textbf{0.757} & \cellcolor{green!15}\textbf{0.824} & \cellcolor{green!15}\textbf{0.762} & \cellcolor{green!15}\textbf{0.742} \\
  3 & Chronos-2-Small (Amazon) & 0.589 & 0.607 & 0.598 & 0.599 \\
  4 & TimesFM 200M (Google) & 0.716 & 0.775 & 0.722 & 0.704 \\
  5 & Chronos-Bolt-Base (Amazon) & 0.385 & 0.548 & 0.399 & 0.384 \\
  6 & Chronos-Bolt-Tiny (Amazon) & 0.417 & 0.575 & 0.430 & 0.411 \\
  7 & Prophet (Meta) & 0.657 & 0.690 & 0.665 & 0.670 \\
  8 & Chronos-Bolt-Small (Amazon) & 0.415 & 0.562 & 0.428 & 0.415 \\
  9 & Chronos-Bolt-Mini (Amazon) & 0.397 & 0.557 & 0.410 & 0.399 \\
  10 & TiRex (NX-AI) & 0.412 & 0.556 & 0.425 & 0.391 \\
  11 & TiRex 1.1 (NX-AI) & 0.437 & 0.577 & 0.449 & 0.414 \\
  12 & TimesFM 2.5 (Google) & 0.381 & 0.529 & 0.395 & 0.371 \\
  13 & PatchTST-FM (IBM) & 0.338 & 0.491 & 0.352 & 0.369 \\
  14 & Moirai 2.0 (Salesforce) & 0.389 & 0.551 & 0.402 & 0.366 \\
  15 & Kairos-50M (Foundation Model Research) & 0.387 & 0.521 & 0.400 & 0.378 \\
  16 & ARIMA (Statsmodels) & 0.607 & 0.622 & 0.607 & 0.611 \\
  17 & Kairos-10M (Foundation Model Research) & 0.366 & 0.509 & 0.380 & 0.356 \\
  18 & TimesFM 500M (Google) & 0.341 & 0.491 & 0.356 & 0.333 \\
  19 & Moirai-Small (Salesforce) & 0.296 & 0.432 & 0.311 & 0.318 \\
  20 & Kairos-23M (Foundation Model Research) & 0.338 & 0.478 & 0.353 & 0.328 \\
  21 & Chronos-Mini (Amazon) & 0.360 & 0.538 & 0.374 & 0.346 \\
  22 & PatchTSMixer (IBM) & 0.167 & 0.371 & 0.183 & 0.226 \\
  23 & Granite FlowState (IBM) & 0.345 & 0.478 & 0.359 & 0.335 \\
  24 & TOTO (Datadog) & 0.106 & 0.255 & 0.125 & -0.099 \\
  25 & Chronos-Tiny (Amazon) & 0.329 & 0.465 & 0.344 & 0.309 \\
  26 & Chronos-Base (Amazon) & 0.240 & 0.405 & 0.256 & 0.269 \\
  27 & Exponential Smoothing (Statsmodels) & 0.023 & 0.202 & 0.047 & 0.084 \\
  28 & Chronos-Small (Amazon) & 0.275 & 0.447 & 0.291 & 0.275 \\
  29 & Moirai-Base (Salesforce) & 0.187 & 0.301 & 0.205 & 0.220 \\
  30 & LSTM (Nixtla) & 0.043 & 0.319 & 0.068 & -0.015 \\
  31 & Croston (Statsmodels) & 0.125 & 0.238 & 0.153 & 0.184 \\
  32 & TTM R2 (IBM) & -0.285 & 0.066 & -0.257 & -0.151 \\
  33 & TTM R1 (IBM) & -0.285 & 0.068 & -0.257 & -0.151 \\
  34 & MOMENT-Large (CMU) & -0.286 & 0.056 & -0.258 & -0.153 \\
  35 & iTransformer (THUML) & 0.162 & 0.186 & 0.181 & 0.216 \\
  36 & Random Forest (scikit-learn) & -0.252 & -0.018 & -0.225 & -0.139 \\
  37 & Time-MoE-200M (Shanghai AI Lab) & -0.998 & -0.499 & -0.955 & -0.775 \\
  38 & Time-MoE-50M (Shanghai AI Lab) & -0.998 & -0.499 & -0.955 & -0.775 \\
  39 & SVR (scikit-learn) & -0.207 & -0.020 & -0.180 & -0.155 \\
  40 & MOMENT-Small (CMU) & -0.302 & -0.047 & -0.274 & -0.177 \\
  41 & N-BEATS (Nixtla) & -0.080 & -0.081 & -0.041 & -0.091 \\
  42 & MOMENT-Base (CMU) & -0.299 & -0.022 & -0.270 & -0.168 \\
  43 & Seasonal Naive (Sktime) & 0.000 & 0.000 & 0.000 & 0.000 \\
  44 & PatchTST-Granite (IBM) & -0.219 & -0.112 & -0.193 & -0.063 \\
  45 & N-HiTS (Nixtla) & -12.563 & -14.671 & -12.588 & -12.257 \\
  46 & Theta (Sktime) & -36.257 & -39.329 & -35.930 & -32.343 \\
  47 & Sundial (THUML) & -94.735 & -94.791 & -94.822 & -98.339 \\
\end{longtable}
\endgroup

\newpage
\section{Forecasters}
\label{sec_app:models}

\begin{table}[!h]
\centering
\caption{Summary of forecasters included in TempusBench.}
\label{tab:models_summary}
\begin{tabularx}{\textwidth}{@{} >{\bfseries}l l L @{}}
\toprule
\textbf{Category} & \textbf{Included Models} & \textbf{Core Characteristics} \\
\midrule

Foundation Models &
\parbox[t]{4cm}{\raggedright \emph{Moirai, Moirai-MoE, TimesFM, TimesFM-2.0, Chronos, Lag-Llama, Toto, MOMENT}} &
\textbf{Paradigm:} Universal, zero-shot/few-shot forecasting. A single large model is pre-trained on massive, diverse datasets and generalizes to new tasks without \samy{re}{dataset-specific }training. \newline
\textbf{Architecture:} Primarily based on Transformers or other deep learning structures like MLP-Mixers. They process raw time series data via patching or novel tokenization schemes. \newline
\textbf{I/O:} Natively handle univariate, multivariate, and covariate data. Often produce probabilistic forecasts. \\
\addlinespace

Classic Machine Learning &
\parbox[t]{4cm}{\raggedright \emph{LSTM, Random Forest (RF), XGBoost, SVR, TabPFN-TS, Tiny Time Mixers (TTM)}} &
\textbf{Paradigm:} Supervised learning models trained per-dataset. They excel at capturing complex, non-linear relationships but require specific training for each task. \newline
\textbf{Architecture:} Diverse. E.g., RNNs (for sequence memory), Tree Ensembles (for interaction effects), and Kernel Methods. \newline
\textbf{I/O:} Typically require explicit feature (e.g., lags, calendar variables) engineering to create a tabular format. Most often produce point forecasts. \\
\addlinespace

Statistical \& Decomposable &
\parbox[t]{4cm}{\raggedright \emph{ARIMA, Holt-Winters, Prophet, Theta Method, Croston's Method, Seasonal Naive}} &
\textbf{Paradigm:} Assume the time series is generated by an underlying statistical process or can be decomposed into simpler, interpretable components like trend and seasonality. \newline
\textbf{Architecture:} An explicit mathematical formula is fit directly to an individual time series. \newline
\textbf{I/O:} Highly interpretable point forecasts. Often specialized for particular data patterns (e.g., intermittency with \emph{Croston's}). \\

\bottomrule
\end{tabularx}
\end{table}

\begin{sidewaystable}[htbp]
\centering
\small %
\caption{Comparative Overview of Forecasting Models}
\label{tab:model_comparison}
\begin{tabularx}{\textwidth}{@{}LLLL@{}}
\toprule
\textbf{Feature} & \textbf{Group 1: Foundation Models} & \textbf{Group 2: Machine Learning Models} & \textbf{Group 3: Classical \& Statistical Models} \\
\midrule

\textbf{Models Included} &
\begin{itemize}[leftmargin=*, topsep=0pt, partopsep=0pt, itemsep=0pt]
    \item \textit{Moirai / Moirai-MoE}
    \item \textit{TimesFM / TimesFM-2.0}
    \item \textit{Chronos}
    \item \textit{Lag-Llama}
    \item \textit{Toto}
    \item \textit{MOMENT}
\end{itemize} &
\begin{itemize}[leftmargin=*, topsep=0pt, partopsep=0pt, itemsep=0pt]
    \item \textit{LSTM}
    \item \textit{Random Forest (RF)}
    \item \textit{XGBoost}
    \item \textit{SVR}
    \item \textit{TabPFN-TS}
    \item \textit{Tiny Time Mixers (TTM)}
\end{itemize} &
\begin{itemize}[leftmargin=*, topsep=0pt, partopsep=0pt, itemsep=0pt]
    \item \textit{ARIMA}
    \item \textit{Holt-Winters}
    \item \textit{Prophet}
    \item \textit{Theta Method}
    \item \textit{Croston's Method}
    \item \textit{Seasonal Naive}
\end{itemize} \\
\addlinespace

\textbf{Core Paradigm} &
Large, pre-trained models designed for universal, zero-shot forecasting. They learn general time-series patterns from massive, diverse datasets. &
Models trained for a specific forecasting task, often relying on feature engineering. They leverage distinct architectures (e.g., recurrence, ensembles) rather than massive pre-training. &
Model-based approaches assuming an underlying stochastic process or decomposable structure. Parameters are estimated directly from the target time series. \\
\addlinespace

\textbf{Architecture} &
Primarily Transformer-based (Encoder, Decoder, or both). Innovations include MoE layers, residual forecasting, and specialized attention mechanisms. &
Diverse architectures: MLP-Mixer (\textit{TTM}), RNN (\textit{LSTM}), Tabular-Transformer (\textit{TabPFN-TS}), Tree Ensembles (\textit{RF, XGBoost}), and Kernel-based (\textit{SVR}). &
Mathematical formulations: State-space models (\textit{ARIMA, Holt-Winters}), decomposable additive models (\textit{Prophet, Theta}), and simple heuristics (\textit{Croston's, Seasonal Naive}). \\
\addlinespace

\textbf{Input Handling} &
Process raw time series, typically via patching (\textit{Moirai, TimesFM}), lag-based tokenization (\textit{Lag-Llama}), or value quantization (\textit{Chronos}). Can natively handle uni/multivariate series. &
Generally require explicit feature (e.g., lags, calendar variables) engineering to create a tabular dataset (\textit{RF, XGBoost, SVR}). \textit{LSTM} and \textit{TTM} process raw sequences. &
Operate directly on the univariate time series. May require stationarity (\textit{ARIMA}) or be specialized for patterns like seasonality (\textit{Holt-Winters}) or intermittency (\textit{Croston's}). \\
\addlinespace

\textbf{Output Type} &
Mostly probabilistic, predicting the parameters of a flexible distribution. Point forecasts are derived from the distribution (e.g., median). &
Primarily point forecasts. \textit{TabPFN-TS} is a notable exception, providing a probabilistic output by approximating the posterior. &
Primarily point forecasts. \textit{Prophet} is an exception, generating uncertainty intervals. Probabilistic versions exist but are not standard. \\
\addlinespace

\textbf{Key Trait} &
Powerful zero-shot/few-shot performance. High model capacity and ability to generalize across domains without dataset-specific training. &
Model-specific strengths: computational efficiency (\textit{TTM}), modeling long-term dependencies (\textit{LSTM}), and capturing complex non-linear interactions (\textit{RF, XGBoost}). &
High interpretability and strong statistical foundations. Often specialized and highly efficient for specific data patterns (e.g., trend, seasonality, intermittency). \\

\bottomrule
\end{tabularx}
\end{sidewaystable}

In this section, we summarize the forecasting models which have been included in \tempusbench{}. We summarize all models in \Cref{tab:models_summary}, and provide and comparison of TSFMs, machine learning forecasting models, and statistical forecasting models in \Cref{tab:model_comparison}.

\paragraph{Moirai}
Moirai \citep{woo2024unified} is a universal time-series forcasting model developed by Salesforce AI Research, built upon a masked encoder-only Transformer architecture. It is designed as a single, large pre-trained model capable of handling diverse forecasting tasks without dataset-specific retraining. The model is pre-trained on LOTSA, a large-scale archive of over 27 billion observations, enabling it to perform powerful zero-shot forecasting.

\begin{itemize}[leftmargin=*]
    \item \textbf{Input:} Accepts univariate or multivariate time series with an arbitrary number of variates and covariates.
    \item \textbf{Output:} Produces a probabilistic forecast by predicting the parameters of a flexible mixture distribution (composed of Student's t, Negative Binomial, Log-Normal, and low-variance Normal distributions) for the next token in an autoregressive manner.
    \item \textbf{Architecture:} Employs a masked encoder-only Transformer. Its key innovations include:
    \begin{itemize}
        \item \textbf{Multi Patch Size Projection:} Uses different patch sizes to effectively process time series of varying frequencies.
        \item \textbf{Any-variate Attention:} Flattens multivariate series into a single sequence and uses binary attention biases to manage an arbitrary number of variates while maintaining permutation equivariance.
    \end{itemize}
    \item \textbf{Forecasting Type:} A universal, zero-shot, probabilistic forecaster. It can generate point forecasts by taking the median of the predicted distribution.
\end{itemize}

\paragraph{Moirai-MoE}
Moirai-MoE \citep{liu2024moirai} is an advanced version of the Moirai foundation model that integrates a Sparse Mixture of Experts (MoE) architecture. Instead of relying on heuristic-based, frequency-specific projection layers, Moirai-MoE delegates the task of modeling diverse time series patterns to specialized "expert" networks within its Transformer layers. This design allows for automatic, token-level specialization in a data-driven manner, leading to improved accuracy and fewer activated parameters, resulting in faster inference. 

\begin{itemize}[leftmargin=*]
    \item \textbf{Input:} Accepts univariate or multivariate time series with an arbitrary number of variates and covariates.
    \item \textbf{Output:} Produces a probabilistic forecast by predicting the parameters of a flexible mixture distribution (composed of Student's t, Negative Binomial, Log-Normal, and low-variance Normal distributions) for the next token in an autoregressive manner.
    \item \textbf{Architecture:} Employs a decoder-only Transformer that replaces the standard Feed-Forward Network (FFN) layers with MoE layers. Key architectural changes from the original Moirai include:
    \begin{itemize}
        \item \textbf{Mixture of Experts (MoE):} A gating function routes each time series token to a small subset of specialized expert networks, allowing the model to handle diverse patterns at a granular level.
        \item \textbf{Single Projection Layer:} It uses a single input/output projection layer for all time series, removing the dependency on frequency-based heuristics.
    \end{itemize}
    \item \textbf{Forecasting Type:} A universal, zero-shot, probabilistic forecaster that is more accurate and efficient (in terms of activated parameters) than the original Moirai model. It can generate point forecasts by taking the median of the predicted distribution.
\end{itemize}

\paragraph{TimesFM}
TimesFM \citep{das2024decoder} is a time-series foundation model developed by Google Research, designed for zero-shot forecasting. It is based on a decoder-only Transformer architecture and is pretrained on a very large corpus of time-series data, combining both real-world and synthetic sources. The model's key objective is to provide accurate out-of-the-box point forecasts on unseen datasets without requiring any dataset-specific training. 

\begin{itemize}[leftmargin=*]
    \item \textbf{Input:} Accepts a univariate time series context window.
    \item \textbf{Output:} Produces a point forecast for a given prediction horizon.
    \item \textbf{Architecture:} Employs a decoder-only Transformer architecture that processes the time series in patches. Key architectural features include:
    \begin{itemize}
        \item \textbf{Decoder-Only Transformer:} Utilizes a standard decoder-style attention mechanism to autoregressively predict future values patch by patch.
        \item \textbf{Input Patching:} The input time series is segmented into non-overlapping patches, which are then embedded using a residual block of MLPs before being fed to the Transformer.
    \end{itemize}
    \item \textbf{Forecasting Type:} A universal, zero-shot, point forecaster designed primarily for long-horizon forecasting tasks.
\end{itemize}

\paragraph{TimesFM-2.0}
TimesFM-2.0 \citep{das2024decoder} is an improved version of the original foundation model from Google Research. While retaining the same decoder-only Transformer architecture, its key innovation lies in forecasting the residual component of a time series after performing a seasonal-trend decomposition. This approach makes the model significantly more accurate, particularly for time series that exhibit clear trends.

\begin{itemize}[leftmargin=*]
    \item \textbf{Input:} Accepts a univariate time series context window.
    \item \textbf{Output:} Produces a point forecast for a given prediction horizon.
    \item \textbf{Architecture:} Based on the original decoder-only Transformer with input patching. The primary architectural update is its residual forecasting methodology:
    \begin{itemize}
        \item \textbf{Seasonal-Trend Decomposition:} The model first decomposes the input series to separate its trend and seasonal components.
        \item \textbf{Residual Forecasting:} The core Transformer then forecasts the residual (the signal remaining after decomposition). This forecast is added back to the projected trend   to produce the final prediction.
    \end{itemize}
    \item \textbf{Forecasting Type:} A universal, zero-shot, point forecaster with enhanced performance on non-stationary time series compared to its predecessor.
\end{itemize}

\paragraph{Chronos}
Chronos \citep{ansari2024chronos} is a family of pretrained time series models developed by Amazon Science that frames forecasting as a language modeling task. The core idea is to "tokenize" time series values by scaling and quantizing them into a fixed vocabulary. By doing so, standard Transformer-based language model architectures can be trained on sequences of these tokens using a cross-entropy loss, effectively learning the "language" of time series.

\begin{itemize}[leftmargin=*]
    \item \textbf{Input:} Accepts a univariate time series context window.
    \item \textbf{Output:} Produces a probabilistic forecast by generating multiple sample future trajectories. A point forecast can be derived from the median of these samples.
    \item \textbf{Architecture:} Based on standard language model architectures (specifically the T5 encoder-decoder family). Its defining characteristic is its unique data preprocessing pipeline:
    \begin{itemize}
        \item \textbf{Tokenization via Quantization:} The model first applies mean scaling to the input time series. It then quantizes these scaled values into a finite set of discrete tokens, converting the continuous series into a sequence of categorical variables.
        \item \textbf{Language Model Training:} The model is trained to predict the next token in a sequence using a standard cross-entropy loss, analogous to how a language model predicts the next word.
    \end{itemize}
    \item \textbf{Forecasting Type:} A universal, zero-shot, probabilistic forecaster.
\end{itemize}

\paragraph{Lag-Llama} \citep{rasul2023lag}
Lag-Llama is a foundation model for univariate probabilistic time-series forecasting. It is built upon a decoder-only Transformer architecture, similar to LLaMA, and is pretrained on a large, diverse corpus of open-source time-series data. The model's key innovation is its tokenization strategy, which uses lagged values of the time series as input features, allowing it to generalize across different frequencies and domains.

\begin{itemize}[leftmargin=*]
    \item \textbf{Input:} Accepts a univariate time series context window.
    \item \textbf{Output:} Produces a probabilistic forecast by outputting the parameters of a Student's t-distribution for the next time step. Future trajectories are generated autoregressively.
    \item \textbf{Architecture:} Based on a decoder-only Transformer (LLaMA). Its defining characteristic is its input representation:
    \begin{itemize}
        \item \textbf{Tokenization via Lag Features:} Instead of patching, the input token for each time step is a vector composed of lagged values from the time series history (e.g., values from 1, 7, and 14 days prior). These values are augmented with standard date-time features.
        \item \textbf{Value Scaling:} Applies robust scaling (using median and IQR) to normalize the input values and includes the scaling parameters as additional features.
    \end{itemize}
    \item \textbf{Forecasting Type:} A universal, zero-shot/few-shot, probabilistic forecaster.
\end{itemize}

\paragraph{Toto}
Toto (Time Series Optimized Transformer for Observability) \citep{cohen2025time} is a foundation model from Datadog, specifically designed for multivariate time-series forcasting with a focus on observability data \sdeni{}{(i.e., latency, throughput, etc)}. It is built on a decoder-only Transformer architecture and incorporates several novel components to handle the unique challenges of observability data, such as high non-stationarity and heavy-tailed distributions. The model is pretrained on a large and diverse corpus that includes real-world observability data, public datasets, and synthetic data.

\begin{itemize}[leftmargin=*]
    \item \textbf{Input:} Accepts multivariate time series.
    \item \textbf{Output:} Produces a probabilistic forecast by predicting the parameters of a Student's T mixture model.
    \item \textbf{Architecture:} A decoder-only Transformer with several key innovations tailored for observability data:
    \begin{itemize}
        \item \textbf{Patch-based Causal Normalization:} A novel per-patch scaling method that computes normalization statistics from current and past data to handle highly nonstationary series.
        \item \textbf{Proportional Factorized Attention:} An efficient attention mechanism that uses a mix of time-wise and variate-wise attention blocks to judiciously model interactions in high-dimensional multivariate data.
        \item \textbf{Student's T Mixture Model Head:} An output layer that models the predictive distribution as a mixture of Student's T distributions to better capture the complex, heavy-tailed nature of observability metrics.
        \item \textbf{Composite Robust Loss:} A hybrid loss function combining negative log-likelihood with a robust point-wise loss to stabilize training in the presence of outliers. %
    \end{itemize}
    \item \textbf{Forecasting Type:} A universal, zero-shot, probabilistic forecaster for multivariate time series.
\end{itemize}

\paragraph{MOMENT}
MOMENT (Multi-task, Open-source, Foundation Model for Time-series) \citep{goswami2024moment} is a family of open-source foundation models from Carnegie Mellon University designed for general-purpose time series analysis. The models are built on a Transformer encoder architecture and are pretrained on a large, diverse collection of public time series called the "Time Series Pile." A key characteristic of MOMENT is its versatility; it is designed to serve as a building block for a wide range of downstream tasks, including forecasting, classification, anomaly detection, and imputation, often with minimal task-specific fine-tuning.

\begin{itemize}[leftmargin=*]
    \item \textbf{Input:} Accepts a univariate time series of a fixed length. Multivariate time series are handled by treating each channel independently.
    \item \textbf{Output:} Produces a reconstructed version of the input time series. This output can be adapted for various downstream tasks, such as generating forecasts by masking future values or extracting embeddings for classification.
    \item \textbf{Architecture:} A standard Transformer encoder that processes time-series data in patches.
    \begin{itemize}
        \item \textbf{Masked Pre-training:} The model is pretrained using a masked time series prediction task. It learns to reconstruct randomly masked patches of the input time series, enabling it to learn robust representations.
        \item \textbf{Patching:} The input time series is segmented into non-overlapping patches, which are then linearly projected into embeddings for the Transformer.
        \item \textbf{Lightweight Prediction Head:} A simple linear layer is used to reconstruct the time series from the Transformer's output embeddings. This head can be easily replaced or adapted for different downstream tasks.
    \end{itemize}
    \item \textbf{Forecasting Type:} A universal foundation model for general time series analysis. It can be used for zero-shot or few-shot forecasting (point-based), classification, anomaly detection, and imputation.
\end{itemize}

\paragraph{Long Short-Term Memory (LSTM)}
Long Short-Term Memory (LSTM) is a type of Recurrent Neural Network (RNN) architecture specifically designed to learn long-term dependencies in sequential data.
Unlike traditional neural networks, LSTMs have internal mechanisms called "gates" that regulate the flow of information. These gates enable the network to selectively remember or forget information over long periods, making it particularly well-suited for time-series forecasting. \citep{siami2018forecasting}

\begin{itemize}[leftmargin=*]
    \item \textbf{Input:} A time-series dataset.
    \item \textbf{Output:} A point forecast for one or more future time steps.
    \item \textbf{Mathematical Formulation:} The core of an LSTM unit is its cell state, $\wh{\c}_t$, which acts as a memory. The flow of information into and out of the cell is controlled by three gates: the forget gate ($\f_t$), the input gate ($\i_t$), and the output gate ($\o_t$). At each time step $t$, these gates update the cell state and produce a hidden state, $\wh{\h}_t$.
    \begin{align}
        \text{Forget Gate:} \quad \f_t &= \vect[\sigma](\mat[W][f] \cdot [\wh{\h}_{t-1}, \targetts[t]] + \vect[b][f]) \\
        \text{Input Gate:} \quad \i_t &= \vect[\sigma](\mat[W][i] \cdot [\wh{\h}_{t-1}, \targetts[t]] + \vect[b][i]) \\
        \text{Candidate State:} \quad \wt{\c}_t &= \tanh(\mat[W][c] \cdot [\wh{\h}_{t-1}, \targetts[t]] + \vect[b][c]) \\
        \text{Cell State Update:} \quad \wh{\c}_t &= \f_t \odot \wh{\c}_{t-1} + \i_t \odot \wt{\c}_t \\
        \text{Output Gate:} \quad \o_t &= \vect[\sigma](\mat[W][o] \cdot [\wh{\h}_{t-1}, \targetts[t]] + \vect[b][o]) \\
        \text{Hidden State Update:} \quad \wh{\h}_t &= \o_t \odot \tanh(\wh{\c}_t)
    \end{align}
    where $\mat[W]$ and $\vect[b]$ are the weight matrices and bias vectors for each gate, $\vect[\sigma]$ is the sigmoid function, and $\odot$ denotes element-wise multiplication. The final prediction is typically generated by passing the hidden state $\wh{\h}_t$ through a dense output layer.
    \item \textbf{Forecasting Type:} A neural network model for point forecasting that can capture complex non-linear patterns in time-series data.
\end{itemize}

\paragraph{Random Forest}
Random Forest is an ensemble machine learning model that operates by constructing a multitude of decision trees at training time. In time-series forcasting, it is applied as a regression model to a featurized dataset. By fitting numerous trees on various sub-samples of the data and employing randomness in feature selection, it improves predictive accuracy and controls over-fitting. The final prediction is an average of the outputs from all individual trees, making the model robust and capable of capturing complex, non-linear relationships. \citep{lin2017random}

\begin{itemize}[leftmargin=*]
    \item \textbf{Input:} A feature matrix $\mat[X]$ where rows are observations and columns are engineered features (e.g., lags, calendar variables), and a corresponding target vector $\y$.
    \item \textbf{Output:} A point forecast for each input feature vector.
    \item \textbf{Architecture and Formulation:} A Random Forest is an ensemble of $B$ decision trees. Its predictive power comes from two sources of randomness introduced during training:
    \begin{itemize}
        \item \textbf{Bagging (Bootstrap Aggregating):} Each individual tree, $f_b$, is trained on a bootstrap sample (a random sample drawn with replacement) from the original training dataset.
        \item \textbf{Feature Randomness:} When splitting a node in a tree, the algorithm considers only a random subset of the total features, which decorrelates the trees in the forest.
    \end{itemize}
    For a new input feature vector $\x$, the forecast is the average of the predictions from all $B$ trees in the ensemble:
    \begin{equation}
        \wh{\y}(\x) = \frac{1}{B} \sum_{b=1}^{B} f_b(\x)
    \end{equation}
    \item \textbf{Forecasting Type:} An ensemble machine learning model for point forecasting. It is non-parametric and highly effective at modeling non-linear relationships between features and the target variable.
\end{itemize}

\paragraph{XGBoost}
XGBoost (Extreme Gradient Boosting) is a powerful and efficient implementation of the gradient boosting framework. It is an ensemble model that builds decision trees sequentially, where each new tree is trained to correct the errors made by the previous ones. For time-series forcasting, XGBoost is used as a regression model on a featurized dataset, making it highly effective at capturing complex, non-linear relationships between the engineered features (e.g., lags, calendar variables) and the target. \citep{wang2023novel}

\begin{itemize}[leftmargin=*]
    \item \textbf{Input:} A feature matrix $\mat[X]$ and a corresponding target vector $\y$.
    \item \textbf{Output:} A point forecast for each input feature vector.
    \item \textbf{Architecture and Formulation:} XGBoost builds an additive model where the final prediction is the sum of the predictions from $K$ decision trees:
    \begin{equation}
        \wh{\y}_i = \sum_{k=1}^{K} f_k(\x_i)
    \end{equation}
    The trees are added one at a time in a greedy fashion. The $k$-th tree, $f_k$, is chosen to minimize a regularized objective function:
    \begin{equation}
        \mathcal{L}^{(k)} = \sum_{i=1}^{n} l(\y_i, \wh{\y}_i^{(k-1)} + f_k(\x_i)) + \Omega(f_k)
    \end{equation}
    where $l$ is a differentiable loss function, $\wh{\y}_i^{(k-1)}$ is the prediction from the first $k-1$ trees, and $\Omega$ is a regularization term that penalizes the complexity of the tree:
    \begin{equation}
        \Omega(f) = \gamma T + \frac{1}{2}\lambda \sum_{j=1}^{T} w_j^2
    \end{equation}
    Here, $T$ is the number of leaves in the tree, $\vect[w]$ is the vector of scores on the leaves, and $\gamma$ and $\lambda$ are regularization parameters.
    \item \textbf{Forecasting Type:} An ensemble machine learning model for point forecasting, known for its high performance, speed, and regularization capabilities.
\end{itemize}

\paragraph{Support Vector Regression (SVR)}
Support Vector Regression (SVR) is a supervised learning algorithm that extends the principles of Support Vector Machines (SVMs) to regression problems. Instead of finding a hyperplane that separates classes, SVR aims to find a function that deviates from the target values by a value no greater than a specified margin, $\epsilon$, for as many of the training points as possible. It is particularly effective in high-dimensional spaces and is robust to some outliers due to its use of an $\epsilon$-insensitive loss function, which ignores errors within this margin. \citep{zhang2020support}

\begin{itemize}[leftmargin=*]
    \item \textbf{Input:} A feature matrix $\mat[X]$ and a corresponding target vector $\y$.
    \item \textbf{Output:} A point forecast for each input feature vector.
    \item \textbf{Mathematical Formulation:} The goal of SVR is to find a function $f(\x) = \w^T \x + b$ that is as "flat" as possible. This is achieved by minimizing the norm of the weight vector, $||\w||^2$. The optimization problem is formulated to tolerate errors up to a margin $\epsilon$ while penalizing points that fall outside this margin using slack variables $\xi_i$ and $\xi_i^*$. The primal optimization problem is:
    \begin{equation}
        \min_{\w, b, \vect[\xi]} \frac{1}{2} ||\w||^2 + C \sum_{i=1}^{n} (\xi_i + \xi_i^*)
    \end{equation}
    subject to the constraints:
    \begin{align}
        \y_i - (\w^T \x_i + b) &\leq \epsilon + \xi_i \\
        (\w^T \x_i + b) - \y_i &\leq \epsilon + \xi_i^* \\
        \xi_i, \xi_i^* &\geq 0
    \end{align}
    Here, $C$ is a regularization parameter that controls the trade-off between the flatness of the model and the amount up to which deviations larger than $\epsilon$ are tolerated. Non-linear relationships are handled by mapping the data to a higher-dimensional space using a kernel function.
    \item \textbf{Forecasting Type:} A machine learning model for point forecasting that is robust to some outliers and effective in high-dimensional feature spaces.
\end{itemize}

\sdeni{}{
\paragraph{TabPFN}
TabPFN \cite{hoo2025tables} is a prior-fitted network (PFN) for tabular classification and regression. It treats supervised learning on tabular data as in-context learning: rather than fitting parameters to a new dataset at test time, TabPFN amortizes training by meta-learning a posterior predictive distribution over datasets sampled from a structured prior (a causal graphical model over synthetic tabular data). At inference, the entire training set is passed as context via a transformer, and predictions for test points are produced in a single forward pass.
\begin{itemize}
\item \textbf{Input:} A tabular dataset (training split) provided as in-context examples, alongside test rows to be predicted. Each row consists of numerical or categorical features with no assumed temporal structure.
\item \textbf{Output:} A predictive distribution over class labels (classification) or real-valued targets (regression); point predictions are obtained from the distribution mode or mean.
\item \textbf{Architecture:}
\begin{itemize}
\item \textit{Prior:} Training data is synthesized from a Bayesian network prior (structural causal models based on graph and noise distributions, and their transformations), enabling meta-training without training on real-world datasets.
\item \textit{Transformer Backbone:} A standard transformer encodes the concatenated context (training rows with labels) and query (test rows), attending jointly over examples and features to approximate the posterior predictive.
\item \textit{Prediction Head:} A lightweight output head maps transformer representations to a continuous distribution, which are post-hoc discretized if the predictand is discrete.
\end{itemize}
\item \textbf{Forecasting Type:} A zero-shot, amortized tabular predictor; not natively a time-series model, though downstream extensions (e.g., TabPFN-TS) adapt TABPFN to forecasting by featurizing time-series windows into tabular form.
\end{itemize}
}

\paragraph{TabPFN-TS}
TabPFN-TS \citep{hoo2025tables} is a novel approach that adapts TabPFN-v2, a general-purpose tabular foundation model, for time-series forecasting. The core methodology involves recasting the forecasting problem as a tabular regression task. This objective is achieved through lightweight feature engineering on the time index, without relying on lagged values. Notably, the underlying TabPFN-v2 model was pretrained exclusively on synthetic tabular data and has not seen any time-series data.

\begin{itemize}[leftmargin=*]
    \item \textbf{Input:} A univariate time series, which is converted into a feature matrix based on timestamps.
    \item \textbf{Output:} Produces a probabilistic forecast by approximating the posterior predictive distribution for each future time step. Point forecasts can be derived from the mean or median of this distribution.
    \item \textbf{Architecture:} It does not use a time-series-specific architecture. Instead, it relies on:
    \begin{itemize}
        \item \textbf{Feature Engineering:} The time series is transformed into a tabular dataset by creating features from timestamps: e.g., standard calendar features (hour of day, day of week, etc.), automatically detected seasonal features via a Fourier transform, and a simple running index.
        \item \textbf{TabPFN-v2 Model:} The generated tabular data is fed into the pretrained TabPFN-v2 model, which performs the regression task to predict future values.
    \end{itemize}
    \item \textbf{Forecasting Type:} A universal, zero-shot, probabilistic forecaster.
\end{itemize}

\paragraph{Tiny Time Mixers (TTM)}
Tiny Time Mixers (TTM) \citep{ekambaram2024tiny} is a family of lightweight pre-trained models from IBM Research, based on the efficient TSMixer architecture. In contrast to large, LLM-based approaches, TTMs are designed to be extremely small (<1M parameters) and fast, while still providing strong zero-shot and few-shot forecasting performance. The models are pre-trained exclusively on a large corpus of public time-series datasets, making them a highly efficient alternative for universal forecasting.

\begin{itemize}[leftmargin=*]
    \item \textbf{Input:} Accepts univariate or multivariate time series, with optional support for exogenous variables during the fine-tuning stage.
    \item \textbf{Output:} Produces a point forecast for a given prediction horizon.
    \item \textbf{Architecture:} Based on the MLP-Mixer architecture. The model is pre-trained in a channel-independent manner and uses a multi-level structure to handle diverse data and tasks.
    \begin{itemize}
        \item \textbf{TSMixer Backbone:} The core of the model uses simple MLP blocks for temporal and feature mixing, avoiding the computational overhead of Transformer-based attention.
        \item \textbf{Multi-Resolution Pre-training:} Employs several novel techniques to handle heterogeneous datasets, including adaptive patching (using different patch configurations at different layers) and data augmentation via downsampling (creating additional training examples by resampling at coarser time resolutions.).
        \item \textbf{Multi-level Modeling:} Uses a frozen pre-trained backbone and a smaller, fine-tunable decoder, which can incorporate channel-mixing and an \textbf{exogenous mixer} to fuse external signals for target-specific tasks.
    \end{itemize}
    \item \textbf{Forecasting Type:} A universal, zero-shot/few-shot, point forecaster, notable for its small size and computational efficiency.
\end{itemize}

\paragraph{ARIMA}
The Autoregressive Integrated Moving Average (ARIMA) \citep{siami2018forecasting} model is a class of statistical models for analyzing and forecasting time-series data. It is a generalization of the simpler Autoregressive Moving Average (ARMA) model that can be applied to non-stationary time series. The model's name reflects its three core components: Autoregression (AR), Integrated (I), and Moving Average (MA). These components capture the key temporal structures within the data, such as dependencies on past observations and past forecast errors.

\begin{itemize}[leftmargin=*]
    \item \textbf{Input:} A univariate time series.
    \item \textbf{Output:} A point forecast for future time steps. While classical ARIMA produces point forecasts, probabilistic forecasts can be generated by assuming a distribution for the error term.
    \item \textbf{Mathematical Formulation:} An ARIMA$(p, d, q)$ model is defined by three parameters: the order of the autoregressive component ($p$), the degree of differencing ($d$), and the order of the moving average component ($q$). The model assumes that the differenced time series, $\wt{\y}_t = (1-B)^d \targetts[t]$, is stationary, where $B$ is the backshift operator. The formulation for the stationary series $\wt{\y}_t$ is:
    \begin{equation}
        \wt{\y}_t = c + \sum_{i=1}^{p} \vect[\phi][i] \wt{\y}_{t-i} + \sum_{j=1}^{q} \vect[\theta][j] \vect[\epsilon][t-j] + \vect[\epsilon][t]
    \end{equation}
    where:
    \begin{itemize}
        \item $p$ is the autoregressive order, representing the number of lagged observations included in the model.
        \item $d$ is the degree of differencing, representing the number of times the raw observations are differenced to achieve stationarity.
        \item $q$ is the moving average order, representing the size of the moving average window applied to past forecast errors.
        \item $\vect[\phi]$ is the vector of autoregressive coefficients.
        \item $\vect[\theta]$ is the vector of moving average coefficients.
        \item $c$ is a constant term.
        \item $\vect[\epsilon][t]$ is the white noise error term at time $t$, typically assumed to be drawn from a Gaussian distribution with zero mean.
    \end{itemize}
    \item \textbf{Forecasting Type:} A statistical model that provides point forecasts. It is often used as a baseline in forecasting tasks. Seasonal variations can be included by using a Seasonal ARIMA (SARIMA) model.
\end{itemize}

\paragraph{Holt-Winters Exponential Smoothing}
Holt-Winters is an extension of exponential smoothing that explicitly models trend and seasonality. It is a widely used statistical method for forecasting time-series data that exhibit these components. The method operates by applying exponential smoothing to three components: the level, the trend, and the seasonality. There are two primary variations of the model, additive and multiplicative, which differ in how they incorporate the seasonal component. \cite{kalekar2004time}

\begin{itemize}[leftmargin=*]
    \item \textbf{Input:} A univariate time series with trend and seasonality.
    \item \textbf{Output:} A point forecast for future time steps.
    \item \textbf{Mathematical Formulation:} The model provides separate updating equations for the level ($\wh{l}_t$), trend ($\wh{b}_t$), and seasonal ($\wh{s}_t$) components, using smoothing parameters $\alpha$, $\beta$, and $\gamma$, respectively. Let $L$ be the length of the seasonal period.
    
    \textbf{Additive Method:} Used when the seasonal variation is roughly constant throughout the series.
    \begin{align}
        \text{Level:} \quad \wh{l}_t &= \alpha(\targetts[t] - \wh{s}_{t-L}) + (1-\alpha)(\wh{l}_{t-1} + \wh{b}_{t-1}) \\
        \text{Trend:} \quad \wh{b}_t &= \beta(\wh{l}_t - \wh{l}_{t-1}) + (1-\beta)\wh{b}_{t-1} \\
        \text{Seasonality:} \quad \wh{s}_t &= \gamma(\targetts[t] - \wh{l}_t) + (1-\gamma)\wh{s}_{t-L}
    \end{align}
    The forecast for $h$ steps ahead from time $t$, denoted $\wh{\y}_{t+h|t}$, is:
    \begin{equation}
        \wh{\y}_{t+h|t} = \wh{l}_t + h\wh{b}_t + \wh{s}_{t-L+h_L^+} \quad \text{where } h_L^+ = \lfloor(h-1) \hspace{-2.5mm} \mod L\rfloor + 1
    \end{equation}

    \textbf{Multiplicative Method:} Used when the seasonal variation changes in proportion to the level of the series.
    \begin{align}
        \text{Level:} \quad \wh{l}_t &= \alpha\left(\frac{\targetts[t]}{\wh{s}_{t-L}}\right) + (1-\alpha)(\wh{l}_{t-1} + \wh{b}_{t-1}) \\
        \text{Trend:} \quad \wh{b}_t &= \beta(\wh{l}_t - \wh{l}_{t-1}) + (1-\beta)\wh{b}_{t-1} \\
        \text{Seasonality:} \quad \wh{s}_t &= \gamma\left(\frac{\targetts[t]}{\wh{l}_t}\right) + (1-\gamma)\wh{s}_{t-L}
    \end{align}
    The forecast for $h$ steps ahead is given by:
    \begin{equation}
        \wh{\y}_{t+h|t} = (\wh{l}_t + h\wh{b}_t) \wh{s}_{t-L+h_L^+} \quad \text{where } h_L^+ = \lfloor(h-1) \hspace{-2.5mm} \mod L\rfloor + 1
    \end{equation}
    
    \item \textbf{Forecasting Type:} A statistical model for point forecasting that can handle various combinations of trend and seasonality.
\end{itemize}

\paragraph{Prophet}
Prophet is a forecasting procedure developed by Meta, based on a decomposable time series model. It is designed to be robust to missing data and shifts in the trend, and it typically handles holidays and seasonal effects well. The model fits an additive model with components for trend, seasonality, and holidays. \citep{anand2024comparative}

\begin{itemize}[leftmargin=*]
    \item \textbf{Input:} A univariate time series with timestamps.
    \item \textbf{Output:} A point forecast, along with uncertainty intervals.
    \item \textbf{Mathematical Formulation:} The Prophet model is specified as a sum of three components:
    \begin{equation}
        \targetts[t] = g(t) + s(t) + h(t) + \vect[\epsilon][t]
    \end{equation}
    where:
    \begin{itemize}
        \item $g(t)$ is the trend component, which is modeled as either a piecewise linear or logistic growth function. This allows the model to capture non-periodic changes in the time series.
        \item $s(t)$ is the seasonality component, which models periodic changes (e.g., yearly, weekly, daily). It is approximated by a Fourier series:
        \begin{equation}
            s(t) = \sum_{n=1}^{N} \left( a_n \cos\left(\frac{2\pi n t}{P}\right) + b_n \sin\left(\frac{2\pi n t}{P}\right) \right)
        \end{equation}
        where $P$ is the period of the seasonality (e.g., 365.25 for yearly).
        \item $h(t)$ is the holiday component, which represents the effects of holidays and special events. It is modeled as a sum of indicator functions for each holiday.
        \item $\vect[\epsilon][t]$ is the error term, assumed to be normally distributed white noise.
    \end{itemize}
    \item \textbf{Forecasting Type:} A decomposable statistical model for point and probabilistic forecasting, particularly effective for business time series with strong seasonal patterns and holiday effects.
\end{itemize}

\paragraph{Theta Method}
The Theta method is a statistical forecasting technique based on the concept of decomposition. It models a time series by breaking it down into two components, or "theta lines." The first line represents the long-term trend of the data, while the second line is constructed to capture the short-term dynamics by modifying   the curvature of the original series \sdeni{}{(i.e., by computing a linear function of the second difference of the time-series))}. These two lines are forecasted independently and then combined to produce the final forecast. The standard Theta model has been shown to be equivalent to Simple Exponential Smoothing with a drift term. \citep{thomakos2015forecasting}

\begin{itemize}[leftmargin=*]
    \item \textbf{Input:} A univariate time series.
    \item \textbf{Output:} A point forecast for future time steps.
    \item \textbf{Mathematical Formulation:} The method decomposes the original time series, $\targetts[t]$, into two theta lines.
    \begin{itemize}
        \item \textbf{Line 1 (Trend Component):} This line is the simple linear trend fit to the data, which is found by ordinary least squares regression:
        \begin{equation}
            \wt{\y}_t^{(1)} = \wh{a} + \wh{b}t
        \end{equation}
        This line is extrapolated linearly to produce a forecast.
        
        \item \textbf{Line 2 (Short-term Component):} This line is constructed by modifying the original series with a coefficient $\theta$. A common and effective choice is $\theta=2$, which doubles the local curvatures of the series. This modified series, $\wt{\y}_t^{(2)}$, is then forecast using Simple Exponential Smoothing (SES).
    \end{itemize}
    The final forecast, $\wh{\y}_{t+h}$, is a simple average of the forecasts from the two lines:
    \begin{equation}
        \wh{\y}_{t+h} = \frac{1}{2} \left( \wh{\y}_{t+h}^{(1)} + \wh{\y}_{t+h}^{(2)} \right)
    \end{equation}
    \item \textbf{Forecasting Type:} A statistical decomposition model for point forecasting, often used as a strong baseline for its simplicity and performance.
\end{itemize}

\paragraph{Croston's Method}
Croston's method \citep{willemain1994forecasting} is a forecasting technique specifically designed for intermittent demand time series, which are characterized by sporadic, non-zero values interspersed with periods of zero demand. The method decomposes the original time series into two separate components: the magnitude of the non-zero demand and the time interval between consecutive demands. By forecasting these two components separately using Simple Exponential Smoothing and then combining them, the model provides a more accurate estimate of the mean demand per period compared to standard smoothing methods, which can be biased when applied to intermittent data.

\begin{itemize}[leftmargin=*]
    \item \textbf{Input:} A univariate time series with intermittent demand.
    \item \textbf{Output:} A point forecast for the average demand per period.
    \item \textbf{Mathematical Formulation:} The method maintains and updates two estimates: one for the non-zero demand size ($\wh{\z}$) and one for the interval between demands ($\wh{\p}$). Let $\alpha$ be a smoothing parameter, let $\targetts[t]$ be the demand at time $t$, and let $q$ be the time elapsed since the last demand. An update occurs only when a non-zero demand is observed ($\targetts[t] > 0$):
    \begin{align}
        \wh{\z}_t &= \wh{\z}_{t-1} + \alpha (\targetts[t] - \wh{\z}_{t-1}) \\
        \wh{\p}_t &= \wh{\p}_{t-1} + \alpha (q - \wh{\p}_{t-1})
    \end{align}
    If demand at time $t$ is zero, the estimates are not updated ($\wh{\z}_t = \wh{\z}_{t-1}$, $\wh{\p}_t = \wh{\p}_{t-1}$) and the interval counter $q$ is incremented. After a demand occurs, $q$ is reset to 1. The final forecast for the mean demand per period, $\wh{\y}_t$, is the ratio of the two smoothed components:
    \begin{equation}
        \wh{\y}_t = \frac{\wh{\z}_t}{\wh{\p}_t}
    \end{equation}
    \item \textbf{Forecasting Type:} A statistical model for point forecasting, specialized for intermittent or "lumpy" demand patterns.
\end{itemize}

\paragraph{Seasonal Naive}
The Seasonal Naive model is a simple yet effective baseline method for forecasting time series with a strong seasonal component. Its core principle is that the forecast for a future period is equal to the last observed value from the same season. For example, as long as the season has not changed, the forecast for this Monday would be the value from last Monday. Despite its simplicity, it serves as a crucial benchmark for more complex models. \citep{wang2023novel}

\begin{itemize}[leftmargin=*]
    \item \textbf{Input:} A univariate time series with a known seasonal period.
    \item \textbf{Output:} A point forecast for future time steps.
    \item \textbf{Mathematical Formulation:} The forecast for $h$ steps ahead from time $t$, denoted $\wh{\y}_{t+h|t}$, is the last observed value from the corresponding season. Let $L$ be the seasonal period (e.g., $L=7$ for daily data with weekly seasonality). The forecast is given by:
    \begin{equation}
        \wh{\y}_{t+h|t} = \y_{t+h-L \cdot k}
    \end{equation}
    where $k = \lceil h/L \rceil$ is an integer that ensures the lagged time index refers to the most recent observation from the target season. For a one-season-ahead forecast ($h=L$), this simplifies to $\wh{\y}_{t+L|t} = \y_t$.
    \item \textbf{Forecasting Type:} A simple statistical baseline for seasonal point forecasting.
\end{itemize}

\newpage
\section{Benchmark Tasks included in \tempusbench{}}
\label{sec_app:benchmarks}

This section describes the datasets for each benchmark task, as summarized in \Cref{tab:task_categories}.

\textit{Notation.} $l$: context window; $h$: forecast horizon (from each task's \texttt{task.yaml}). $n$: number of target variables; $m$: number of covariate variables (CSV rows with \texttt{variable\_type} \texttt{target} vs.\ \texttt{covariate}). \textbf{Tot.\ pts.}: total number of scalar observations $T(n+m)$ for the full stored series, with $T$ the common time length.

\smallskip
\begingroup
\footnotesize
\setlength{\tabcolsep}{2.6pt}
\renewcommand{\arraystretch}{0.88}
\setlength{\LTpre}{6pt}
\setlength{\LTpost}{6pt}

\begin{longtable}{@{}>{\raggedright\arraybackslash}p{3.35cm}>{\raggedright\arraybackslash}p{4.95cm}cccccc@{}}
\caption{Summary of \textbf{univariate} benchmark tasks}
\label{tab:benchmark_summary_uni}\\
\toprule
\textbf{Category} & \textbf{Task} & \(l\) & \(h\) & \(n\) & \(m\) & \textbf{Tot.\ pts.} \\
\midrule
\endfirsthead
\multicolumn{7}{c}{\tablename~\thetable{} --- \textit{continued from previous page}} \\
\toprule
\textbf{Category} & \textbf{Task} & \(l\) & \(h\) & \(n\) & \(m\) & \textbf{Tot.\ pts.} \\
\midrule
\endhead
\midrule
\multicolumn{7}{r}{\textit{Continued on next page}} \\
\endfoot
\endlastfoot
\multicolumn{7}{@{}l@{}}{\textbf{Trend}} \\
\midrule
Non-stationary & Software job postings \citep{fred2025indeed} & 512 & 64 & 1 & 0 & 1{,}827 \\
\midrule
\multicolumn{7}{@{}l@{}}{\textbf{Decomposition}} \\
\midrule
Additive & Synthetic additive (\Cref{ssec:cyclic_additive}) & 1024 & 64 & 1 & 0 & 3{,}000 \\
Multiplicative & Synthetic multiplicative (\Cref{ssec:cyclic_multiplicative}) & 1024 & 64 & 1 & 0 & 3{,}000 \\
\midrule
\multicolumn{7}{@{}l@{}}{\textbf{Frequency}} \\
\midrule
Days & Coinbase Litecoin \citep{fred2025cbltcusd} & 512 & 64 & 1 & 0 & 1{,}827 \\
Months & Inventories/sales \citep{fred2025mnfctrirsa} & 64 & 30 & 1 & 0 & 402 \\
Quarters & German house prices \citep{fred2025qder628bis} & 32 & 17 & 1 & 0 & 221 \\
Weeks & Federal funds rate \citep{fred2025ff} & 1024 & 64 & 1 & 0 & 3{,}713 \\
Years & Personal consumption \citep{fred2025pce} & 32 & 5 & 1 & 0 & 96 \\
\midrule
\multicolumn{7}{@{}l@{}}{\textbf{Seasonality}} \\
\midrule
Periodic & Synthetic cyclic & 1024 & 64 & 1 & 0 & 3{,}000 \\
Quasiperiodic & Synthetic non-stationary & 1024 & 64 & 1 & 0 & 3{,}000 \\
\midrule
\multicolumn{7}{@{}l@{}}{\textbf{Domain}} \\
\midrule
Climate & Delhi climate \citep{sumanthvrao2021dailyclimate} & 512 & 64 & 1 & 0 & 1{,}462 \\
Economics/Finance & Coinbase Litecoin \citep{fred2025cbltcusd} & 512 & 64 & 1 & 0 & 1{,}827 \\
Energy & Room SplitSmart \citep{bitspilani2024splitsmart} & 128 & 39 & 1 & 0 & 561 \\
Healthcare & Employees health care \citep{fred2025ces6562} & 64 & 33 & 1 & 0 & 427 \\
Manufacturing & Inventories/sales \citep{fred2025mnfctrirsa} & 64 & 30 & 1 & 0 & 402 \\
Nature & Soil monitoring \citep{noeyislearning2024soil} & 32 & 8 & 1 & 0 & 679 \\
Sales & German house prices \citep{fred2025qder628bis} & 32 & 17 & 1 & 0 & 221 \\
Software & Software job postings \citep{fred2025indeed} & 512 & 64 & 1 & 0 & 1{,}827 \\
Transport & Madrid BEN pollution \citep{banuelos-gimeno2024initial} & 2048 & 64 & 1 & 0 & 172{,}622 \\
Web & Web traffic \citep{raminhuseyn2024webtraffic} & 1024 & 64 & 1 & 0 & 2{,}793 \\
\midrule
\multicolumn{7}{@{}l@{}}{\textbf{Data sparsity}} \\
\midrule
Dense & Chickenpox \citep{uci2021chickenpox} & 128 & 35 & 1 & 0 & 522 \\
Sparse & Patient chart \citep{johnson2019mimic} & 12 & 4 & 1 & 0 & 8{,}093 \\
\midrule
\multicolumn{7}{@{}l@{}}{\textbf{Value type}} \\
\midrule
Binary & Absenteeism \citep{martiniano2012absenteeism} & 128 & 55 & 1 & 0 & 740 \\
Categorical & Online retail \citep{chen2015online} & 2048 & 64 & 1 & 0 & 541{,}909 \\
Count & Occupancy \citep{singh2018room} & 450 & 64 & 1 & 0 & 10{,}129 \\
\bottomrule
\end{longtable}

\smallskip

\begin{longtable}{@{}>{\raggedright\arraybackslash}p{3.35cm}>{\raggedright\arraybackslash}p{4.95cm}cccccc@{}}
\caption{Summary of \textbf{multivariate} benchmark tasks}
\label{tab:benchmark_summary_multi}\\
\toprule
\textbf{Category} & \textbf{Task} & \(l\) & \(h\) & \(n\) & \(m\) & \textbf{Tot.\ pts.} \\
\midrule
\endfirsthead
\multicolumn{7}{c}{\tablename~\thetable{} --- \textit{continued from previous page}} \\
\toprule
\textbf{Category} & \textbf{Task} & \(l\) & \(h\) & \(n\) & \(m\) & \textbf{Tot.\ pts.} \\
\midrule
\endhead
\midrule
\multicolumn{7}{r}{\textit{Continued on next page}} \\
\endfoot
\endlastfoot
\multicolumn{7}{@{}l@{}}{\textbf{Frequency}} \\
\midrule
Days & Gold India \citep{chodavadiya2025goldprice} & 1024 & 64 & 5 & 0 & 20{,}120 \\
Hours & Madrid transport pollution \citep{ignacioqg2022pollution} & 2048 & 64 & 14 & 0 & 2{,}544{,}542 \\
Minutes & U.S.\ stocks2003--24 \citep{deltatrup2024lt} & 2048 & 64 & 6 & 0 & 732{,}660 \\
Minutes & U.S.\ stocks 2003--24 (longest) \citep{deltatrup2024lt} & 2048 & 64 & 6 & 0 & 732{,}660 \\
Months & Airlines baggage \citep{gabrielsantello2023airline} & 32 & 8 & 4 & 0 & 336 \\
Manufacturing & Utah drilling \citep{egi2025utahforge} & 2048 & 64 & 35 & 0 & 338{,}135 \\
\midrule
\multicolumn{7}{@{}l@{}}{\textbf{Seasonality}} \\
\midrule
Periodic & Madrid transport (cyclical) \citep{ignacioqg2022pollution} & 2048 & 64 & 14 & 0 & 2{,}544{,}542 \\
\midrule
\multicolumn{7}{@{}l@{}}{\textbf{Domain}} \\
\midrule
Economics/Finance & Gold India \citep{chodavadiya2025goldprice} & 1024 & 64 & 5 & 0 & 20{,}120 \\
Economics/Finance & Gold India (real) \citep{chodavadiya2025goldprice} & 1024 & 64 & 5 & 0 & 20{,}120 \\
Energy & Room SplitSmart \citep{bitspilani2024splitsmart} & 2048 & 64 & 2 & 0 & 21{,}206 \\
Healthcare & NYC Covid \citep{nyc2025covid} & 512 & 64 & 54 & 0 & 108{,}270 \\
Nature & Soil monitoring \citep{noeyislearning2024soil} & 1024 & 64 & 127 & 0 & 549{,}021 \\
Nature & Soil monitoring (500) \citep{noeyislearning2024soil} & 1024 & 64 & 127 & 0 & 549{,}021 \\
Sales & Airlines baggage \citep{gabrielsantello2023airline} & 32 & 8 & 4 & 0 & 336 \\
Software & Water-network attacks \citep{taormina2018battle} & 512 & 64 & 44 & 0 & 76{,}604 \\
Transport & Airlines baggage (100) \citep{gabrielsantello2023airline} & 32 & 8 & 4 & 0 & 336 \\
Transport & Madrid BEN pollution \citep{banuelos-gimeno2024initial} & 2048 & 64 & 14 & 0 & 2{,}544{,}542 \\
Transport & Madrid BEN pollution (noisy) \citep{ignacioqg2022pollution} & 2048 & 64 & 14 & 0 & 2{,}544{,}542 \\
\midrule
\multicolumn{7}{@{}l@{}}{\textbf{Data sparsity}} \\
\midrule
Dense & Gold India \citep{chodavadiya2025goldprice} & 1024 & 64 & 5 & 0 & 20{,}120 \\
\midrule
\multicolumn{7}{@{}l@{}}{\textbf{Value type}} \\
\midrule
Continuous & Gold India \citep{chodavadiya2025goldprice} & 1024 & 64 & 5 & 0 & 20{,}120 \\
Count & Madrid BEN pollution \citep{banuelos-gimeno2024initial} & 2048 & 64 & 14 & 0 & 2{,}544{,}542 \\
\bottomrule
\end{longtable}

\smallskip

\begin{longtable}{@{}>{\raggedright\arraybackslash}p{3.35cm}>{\raggedright\arraybackslash}p{4.95cm}cccccc@{}}
\caption{Summary of \textbf{covariate} benchmark tasks}
\label{tab:benchmark_summary_co}\\
\toprule
\textbf{Category} & \textbf{Task} & \(l\) & \(h\) & \(n\) & \(m\) & \textbf{Tot.\ pts.} \\
\midrule
\endfirsthead
\multicolumn{7}{c}{\tablename~\thetable{} --- \textit{continued from previous page}} \\
\toprule
\textbf{Category} & \textbf{Task} & \(l\) & \(h\) & \(n\) & \(m\) & \textbf{Tot.\ pts.} \\
\midrule
\endhead
\midrule
\multicolumn{7}{r}{\textit{Continued on next page}} \\
\endfoot
\endlastfoot
\multicolumn{7}{@{}l@{}}{\textbf{Trend}} \\
\midrule
Non-stationary & NIFTY-50 minute covariates (non-stationary) \citep{kaggle_nifty50_minute} & 2048 & 64 & 1 & 4 & 2{,}001{,}355 \\
Noisy data & U.S.\ macro covariates (high-variation panel) \citep{kaggle_us_economy_macro} & 18 & 5 & 1 & 6 & 336 \\
\midrule
\multicolumn{7}{@{}l@{}}{\textbf{Frequency}} \\
\midrule
Days & U.S.\ equities (daily covariates) \citep{kaggle_stock_timeseries_lee} & 2048 & 64 & 1 & 4 & 15{,}095 \\
Hours & SoCal energy (hourly covariates) \citep{kaggle_socal_energy} & 2048 & 64 & 1 & 34 & 1{,}840{,}475 \\
Minutes & NIFTY-50 minute sector covariates \citep{kaggle_nifty50_minute} & 2048 & 64 & 1 & 4 & 2{,}001{,}355 \\
longest & Long weather covariate streams \citep{kaggle_weather_rohitgrewal} & 2048 & 64 & 1 & 4 & 2{,}001{,}355 \\
Months & Monthly FX covariates \citep{kaggle_currency_monthly} & 200 & 16 & 1 & 3 & 1{,}248 \\
Years & U.S.\ macro (annual covariates) \citep{kaggle_us_economy_macro} & 18 & 5 & 1 & 6 & 336 \\
\midrule
\multicolumn{7}{@{}l@{}}{\textbf{Seasonality}} \\
\midrule
Cyclical & Weather covariates (diurnal/seasonal) \citep{kaggle_weather_rohitgrewal} & 2048 & 64 & 1 & 5 & 52{,}704 \\
\midrule
\multicolumn{7}{@{}l@{}}{\textbf{Domain}} \\
\midrule
Climate & Multi-year weather/climate covariates \citep{kaggle_weather_rohitgrewal} & 2048 & 64 & 1 & 5 & 52{,}704 \\
Economics/Finance & U.S.\ equities (econ./finance covs.) \citep{kaggle_stock_timeseries_lee} & 2048 & 64 & 1 & 4 & 15{,}095 \\
real & U.S.\ equities (real-return covs.) \citep{kaggle_stock_timeseries_lee} & 2048 & 64 & 1 & 4 & 15{,}095 \\
Energy & SoCal energy consumption covariates \citep{kaggle_socal_energy} & 2048 & 64 & 1 & 34 & 1{,}840{,}475 \\
Healthcare & NYC Covid (covariate design) \citep{nyc2025covid} & 1536 & 42 & 1 & 53 & 108{,}270 \\
Manufacturing & Construction performance covariates \citep{kaggle_construction_performance} & 2048 & 64 & 1 & 15 & 800{,}000 \\
$\sim$500 & Solar/weather (wide cov.\ field) \citep{kaggle_solar_weather_forecast} & 256 & 22 & 3 & 9 & 6{,}000 \\
$\sim$100 & Solar/weather (moderate field) \citep{kaggle_solar_weather_forecast} & 32 & 6 & 3 & 9 & 1{,}200 \\
Nature & Solar irradiance + weather covariates \citep{kaggle_solar_weather_forecast} & 256 & 64 & 3 & 9 & 12{,}000 \\
Sales & Walmart retail covariates \citep{kaggle_walmart_sales} & 45 & 16 & 45 & 225 & 38{,}610 \\
Software & Cybertec IIoT malware (software telemetry) \citep{kaggle_cybertec_iot_malware} & 2048 & 64 & 1 & 25 & 1{,}177{,}514 \\
Transport & Smart mobility covariates \citep{kaggle_smart_mobility} & 2048 & 64 & 1 & 10 & 55{,}000 \\
\midrule
\multicolumn{7}{@{}l@{}}{\textbf{Data sparsity}} \\
\midrule
Dense & Monthly FX (dense cov.\ panel) \citep{kaggle_currency_monthly} & 200 & 16 & 1 & 3 & 1{,}248 \\
\midrule
\multicolumn{7}{@{}l@{}}{\textbf{Value type}} \\
\midrule
Continuous & U.S.\ equities (continuous covs.) \citep{kaggle_stock_timeseries_lee} & 2048 & 64 & 1 & 4 & 15{,}095 \\
Count & Cybertec IIoT (count covariates) \citep{kaggle_cybertec_iot_malware} & 2048 & 64 & 1 & 25 & 1{,}177{,}514 \\
\bottomrule
\end{longtable}

\smallskip

\endgroup

\subsection{Synthetic Data: Cyclic Seasonality with Additive Trends}
\label{ssec:cyclic_additive}

In this section, we describe how we construct synthetic data with cyclic seasonality using additive trends only.

\subsubsection{Description}
This category of synthetic data models a time series that exhibits both a complex seasonal pattern and a persistent, long-term trend. These data are generated using two related methods. Both start with a foundational signal that combines multi-frequency sinusoids with a linear trend. The second, more complex method builds upon this foundation by introducing an additional, randomized sinusoidal component to the signal.

In both cases, non-negative noise from an exponential distribution is added to the deterministic signal. These datasets are ideal for testing a model's ability to identify and separate periodicities from an underlying linear trend, with the second method providing a more complex seasonal structure.

\subsubsection{Mathematical Formulation}
The generation process for both methods is based on a primary signal, $y_{\text{base}}(t)$, which includes seasonal, trend, and offset components:
\begin{equation}
    y_{\text{base}}(t) = \underbrace{2\sin(t) + 2\cos\left(\frac{t}{2}\right)}_{\text{Seasonality}} + \underbrace{\frac{1}{4}t}_{\text{Trend}} + \underbrace{4}_{\text{Offset}}
\end{equation}

\subsubsection{Method 1: Fixed Additive Trend}
In the first method, the true signal, $y_1(t)$, is simply the base function. The final observed value, $Y_t$, is this signal plus an additive noise term, $\epsilon_t$.
\begin{equation}
    Y_t = y_1(t) + \epsilon_t = y_{\text{base}}(t) + \epsilon_t
\end{equation}

\subsubsection{Method 2: Randomized Additive Trend}
\label{ssec:cyclic_randomized_additive}
The second method introduces additional complexity. For each generated time series, a random frequency parameter, $\alpha$, is sampled once from a continuous uniform distribution:
\begin{equation}
    \alpha \sim U(a, b)
\end{equation}
In the provided code, this range is fixed from $a=0$ to $b=5$. This parameter is used to create an additional sinusoidal component that is added to the base signal. The true signal, $y_2(t)$, is therefore:
\begin{equation}
    y_2(t) = y_{\text{base}}(t) + \sin(\alpha t)
\end{equation}
The final observed value, $Y_t$, is this enhanced signal plus the noise term:
\begin{equation}
    Y_t = y_2(t) + \epsilon_t
\end{equation}

\subsubsection{Noise Model}
\label{ssec:cyclic_additive_noise_model}
In both of the aforementioned methods, the noise term, $\epsilon_t$, is drawn from an exponential distribution with a scale parameter $\beta$:
\begin{equation}
    \epsilon_t \sim \text{Exponential}(\beta)
\end{equation}

\subsubsection{Adjustable Parameters}
\label{ssec:cyclic_additive_adjustable_parameters}
The data generation process is controlled by the following parameters.
\begin{itemize}
    \item \textbf{Number of Points (\texttt{num\_points}, $N$):} This integer parameter sets the total number of data points, defining the length of the time series.

    \item \textbf{Start Time (\texttt{start\_time}, $t_0$):} This parameter defines the initial time value for the series.

    \item \textbf{Noise Scale (\texttt{noise\_std}, $\beta$):} This parameter represents the scale (and mean) of the exponential noise distribution. A larger value of $\beta$ increases the average magnitude of the positive noise added to the base signal.

    \item \textbf{Random Frequency (\texttt{alpha}, $\alpha$):} (Method 2 only) This parameter is not set by the user, but is sampled internally from a uniform distribution $U(0, 5)$ for each generated series. It introduces variability in the seasonal component across different datasets created using the second method.
\end{itemize}

\subsection{Synthetic Data: Cyclic Seasonality with Multiplicative and Additive Trends}
\label{ssec:cyclic_multiplicative}

In this section, we describe how we construct synthetic data with cyclic seasonality based on an amalgamation of additive and multiplicative trends.

\subsubsection{Description}
This category of synthetic data models a time series characterized by a complex interaction of seasonal components and trends. A key feature is a multiplicative trend, where the amplitude of one of the seasonal components grows exponentially over time. This is combined with another stable seasonal component and a linear additive trend.

These data are generated using two related methods. The first method uses a fixed, deterministic signal. The second method introduces additional complexity by adding another sinusoidal component with a randomized frequency to the base signal. In both cases, non-negative noise from an exponential distribution is added. These datasets are particularly useful for testing a model's ability to handle heteroskedasticity, where the variance of the series changes over time, in the presence of other seasonalities and trends.

\subsubsection{Mathematical Formulation}
Both methods are built upon a primary signal, $y_{\text{base}}(t)$, which is a composite of several functions:
\begin{equation}
    y_{\text{base}}(t) = \underbrace{e^{t/100}\sin(t)}_{\text{Multiplicative Seasonality}} + \underbrace{3\cos\left(\frac{t}{2}\right)}_{\text{Additive Seasonality}} + \underbrace{\frac{1}{2}t}_{\text{Linear Trend}}
\end{equation}

\subsubsection{Method 1: Fixed Multiplicative Trend}
In the first method, the true signal, $y_1(t)$, is simply the base function. The final observed value, $Y_t$, is this signal plus an additive noise term, $\epsilon_t$.
\begin{equation}
    Y_t = y_1(t) + \epsilon_t = y_{\text{base}}(t) + \epsilon_t
\end{equation}

\subsubsection{Method 2: Randomized Additive Component}

This method is idential to the Randomized Additive Component in Section~\ref{ssec:cyclic_randomized_additive}.

\subsubsection{Noise Model}
The noise model is identical to the noise model in Section~\ref{ssec:cyclic_additive_noise_model}.

\subsubsection{Adjustable Parameters}
The adjustable parameters are identical to those of Section~\ref{ssec:cyclic_additive_adjustable_parameters}.

\newpage
\section{Benchmark Evaluations}
\label{sec_app:evaluations}

All results for all metrics, tasks, and models are available on the \href{https://smlcrm.com/tempusbench}{TempusBench webpage}.
You can inspect the interactive leaderboard and download a ZIP archive.
That bundle includes, for each task type (univariate, multivariate, covariate), leaderboard, per-metric win-rate, and per-metric skill-score tables in ready-made \LaTeX{} and CSV files under \texttt{latex/} and \texttt{csv/}, which can be used 
as needed. In addition, per-task tables are displayed in the Task Explorer.

\section*{Analysis of Univariate Task Performance}

The strongest overall win rates in that snapshot  are led by \textbf{TiRex} (NX-AI, $\sim$300M parameters) at \textbf{72.6\%}, followed by \textbf{TiRex 1.1} at \textbf{71.3\%}.
\textbf{TimesFM 2.5} (Google, $\sim$200M) is third at \textbf{70.3\%}, then \textbf{Moirai 2.0} (Salesforce, $\sim$12M) at \textbf{67.4\%}, and \textbf{Chronos-2} (Amazon, $\sim$120M) at \textbf{65.8\%}.
Additional foundation checkpoints in the same top tier include \textbf{TimesFM 500M} (64.8\%), \textbf{PatchTST-FM} (IBM, 63.4\%), and compact \textbf{Chronos-Bolt} variants (e.g., Mini63.2\%, Tiny 63.0\%, Base 61.8\%).

On \textbf{pooled averages} of the scalar columns (mean over all task$\times$window rows feeding that model), \textbf{TiRex 1.1} attains the best displayed MAE (0.418), CRPS (0.252), quantile score (0.126), and weighted interval score (0.252).
\textbf{TimesFM 2.5} achieves the best pooled RMSE (0.556).
\textbf{Chronos-Tiny} obtains the lowest pooled MAPE among models in the table for this slice; MAPE is not comparable across tasks without normalization, so task-level exports should be read before drawing fine-grained conclusions.
Pooled MASE can be dominated by a few heavy-tailed tasks; use per-task MASE from the downloadable CSV rather than the column average alone.

\section*{Analysis of Multivariate Task Performance}

\textbf{TiRex} leads win rate at \textbf{82.8\%}, narrowly followed by \textbf{TiRex 1.1} at \textbf{82.0\%}.
\textbf{TimesFM 2.5} is third (\textbf{79.2\%}), \textbf{Chronos-2} fourth (\textbf{77.7\%}), and \textbf{Chronos-2-Small} (40M) fifth (\textbf{74.8\%}).
\textbf{TimesFM 500M}, \textbf{PatchTST-FM}, and \textbf{TabPFN-TS} (Prior Labs, $\sim$1M parameters) occupy the next ranks at \textbf{73.1\%}, \textbf{72.5\%}, and \textbf{66.6\%} respectively---highlighting that a billion-scale parameter gap is not required to stay in the upper win-rate band on this slice.

For pooled CRPS, quantile score, and WIS, \textbf{TiRex 1.1} is best on average; it also leads pooled MAE (0.336) and RMSE (0.548).
\textbf{TimesFM 2.5} achieves the best pooled MAPE and MASE among models in the table for this snapshot.

\subsection*{Comparison of univariate and multivariate results}

As \textbf{TiRex} and \textbf{TiRex 1.1} achieve the top-two win rates on both univariate and multivariate data, with \textbf{TimesFM 2.5} and \textbf{Chronos-2} in the top five for both, suggesting that the best univariate \samy{checkpoints}{models} also scale to \samy{joint}{multivariate} modeling.
Covariate tasks reshuffle leaders, however; Chronos-family models move up.

\section*{Analysis of Covariate Task Performance}

Win-rate order differs on this task than univariate and multivariate: \textbf{Chronos-2} places first at \textbf{79.3\%}, with \textbf{TabPFN-TS} second at \textbf{78.6\%} despite its small footprint.
\textbf{Chronos-2-Small} is third (\textbf{74.8\%}), then \textbf{TimesFM 200M} (\textbf{70.2\%}) and \textbf{Chronos-Bolt-Base} (\textbf{66.0\%}).
\textbf{Prophet} (Meta) achieves a \textbf{65.3\%} win rate, illustrating that classical baselines stay relevant when exogenous regressors are explicit.

On pooled averages, \textbf{Chronos-2} leads CRPS, quantile score, WIS, and RMSE; \textbf{TabPFN-TS} leads MAE and MASE.
\textbf{Chronos-Bolt-Tiny} shows the lowest pooled MAPE in this slice.

\newpage
\section{Additional Related Work}
\label{sec_app:related_works}
Classical time-series forecasting began with statistical models that exploit stochastic structure and domain priors, including ARIMA and its Box–Jenkins methodology \cite{boxjenkins2015}, exponential-smoothing state-space ETS \cite{hyndman2008forecasting}, the Theta method \cite{assimakopoulos2000theta}, and multivariate VAR models \cite{lutkepohl2005new}. Deep learning methods later advanced accuracy and scale by learning nonlinear temporal dependencies from large corpora: DeepAR \cite{flunkert2017deepar}, N-BEATS \cite{oreshkin2020nbeats}, DLinear \cite{zeng2023transformers}, TiDE \cite{das2023tide}, TFT \cite{lim2019tft}, PatchTST \cite{nie2023patchtst}, and iTransformer \cite{liu2023itransformer}. Probabilistic forecasters further model predictive distributions, e.g., diffusion-based TimeGrad \cite{rasul2021timegrad}, score-based CSDI for imputation and forecasting \cite{tashiro2021csdi}, and conditional-flow GRU-NVP \cite{rasul2020grunvp}.

\paragraph{TSFMs.}  Inspired by NLP/vision pretraining, TSFMs train on heterogeneous corpora and evaluate in zero/few-shot settings across domains and horizons. Representative models include Moirai \cite{woo2024moirai}, Chronos \cite{ansari2024chronos}, TimesFM \cite{das2023timesfm}, Lag-Llama \cite{rasul2023lagllama}, Timer \cite{liu2024timer}, UniTS \cite{gao2024units}, TTM (Tiny Time Mixers) \cite{ekambaram2024ttm}, Moment \cite{goswami2024moment}, and multimodal VisionTS \cite{chen2024visionts}. Collectively, they demonstrate strong zero-shot point and probabilistic accuracy on diverse benchmarks while revealing open challenges at long horizons (error accumulation) and at very high frequencies.

\paragraph{Public datasets and repositories.}  Public corpora have underpinned progress from statistical to foundation-model eras. The M-competitions (M3 and M4) provided broad univariate benchmarks across domains and frequencies \cite{makridakis2000m3,makridakis2018m4}, followed by the retail-demand M5 competition \cite{makridakis2022m5}. The Monash Time-Series Forecasting Archive curates a large, standardized repository spanning many domains and sampling granularities \cite{godahewa2021monash}. Large-scale pretraining/evaluation collections include LOTSA (released with Moirai) \cite{woo2024moirai}, the Chronos corpus with in-domain/zero-shot splits \cite{ansari2024chronos}, and the diverse univariate corpus aggregated in Lag-Llama \cite{rasul2023lagllama}. Task-focused collections such as the LTSF suite \cite{zeng2023transformers} (e.g., ETT datasets) and broader benchmarks like TFB \cite{qiu2024tfb} and ProbTS \cite{zhang2023probts} assemble datasets emphasizing horizon length, covariates, and probabilistic outputs.

\paragraph{Evaluation frameworks and benchmarks.}  Tooling and standardized evaluation have evolved in parallel. Practitioner libraries such as Prophet \cite{taylor2018prophet} and sktime \cite{loning2019sktime} offer classical and ML baselines with unified interfaces, while GluonTS \cite{alexandrov2020gluonts} and PyTorchTS \cite{rasul2021pytorchts} provide probabilistic deep-learning pipelines. Benchmarking efforts including LTSF \cite{zeng2023transformers}, BasicTS+ \cite{shao2024exploring} , TFB \cite{qiu2024tfb}, and ProbTS \cite{zhang2023probts} compare statistical, deep, and (in some cases) foundation models, but differ in task taxonomies, splits, and leakage controls. Standardized metrics such as MASE \cite{hyndman2006mase} and CRPS \cite{gneiting2007strictly} enable cross-dataset aggregation of point and probabilistic performance, yet consistent pretraining/evaluation protocols and leakage-free large-scale corpora remain key needs for fair TSFM assessment.

The consequence of these myriad issues is a research environment where it is difficult to distinguish genuine methodological advances from circumstantial performance on a narrow, and potentially contaminated, set of tasks. This is particularly damaging for the development of TSFMs. The significant computational and financial resources required to pre-train these models demand a rigorous, fair, and comprehensive evaluation framework to justify their development and guide future research \citep{thomakos2015forecasting}. The current state of affairs falls short of this standard. Indeed, studies have shown that existing TSFMs, often pre-trained on general-purpose academic datasets, can struggle to generalize to the unique and challenging characteristics of specialized domains like observability data \citep{tamatta2018time}.

The field has thus reached an inflection point. Progress is no longer 
limited by our ability to design novel model architectures, but by our inability to reliably and fairly measure their performance. Recognizing this crisis, recent efforts have focused on creating the next generation of evaluation infrastructure. The development of large-scale, standardized benchmarks such as GIFT-Eval and the domain-specific Benchmark of Observability Metrics (BOOM) represent a direct and necessary response \citep{siami2018forecasting}. These initiatives introduce carefully curated and decontaminated pre-training and evaluation sets, standardized protocols, and data that reflects the complexity of real-world applications. They treat the benchmark not as a mere dataset, but as a carefully designed scientific instrument \citep{anand2024comparative}. A clear and urgent research gap has thus emerged: the critical need for a new, large-scale, and meticulously curated public benchmark to serve as a gold standard for evaluating the next generation of time-series models. Such a contribution is not merely a prerequisite for future research; it provides essential infrastructure required to move the field from an era of fragmented claims to one of robust, reproducible, and generalizable progress \citep{goswami2024moment}, a foundational scientific contribution in its own right.

Contemporary time-series data seldom conform to the idealized assumptions of stationarity and linearity that underpin classical models. Instead, real-world data streams are characterized by a confluence of complex, interacting properties that present formidable modeling challenges \citep{wang2023novel}.
\begin{itemize}
\item \textbf{Non-Linearity:} Perhaps the most fundamental challenge is the prevalence of non-linear relationships. Economic systems, biological processes, and energy grids are governed by complex feedback loops and interactions that cannot be adequately captured by linear models \citep{siami2018forecasting}. Traditional methods like Autoregressive Integrated Moving Average (ARIMA) are, by their construction, limited in their ability to model such non-linear dynamics, which is a primary reason for their performance ceiling on complex, real-world problems \citep{wang2023novel}.

\item \textbf{Multi-Regime Behavior:} Many time series exhibit structural breaks or distinct operational regimes, where the underlying data-generating process changes over time \citep{kalekar2004time}. Examples include the shift between bull and bear markets in financial data or the different performance characteristics of an industrial machine under varying loads and environmental conditions. A single, global model often fails to capture this complex inner structure, leading to significant predictive errors when the system transitions between regimes \citep{haskins2021financial}.

\item \textbf{Intermittency:} As noted previously, intermittent demand patterns are characterized by a high proportion of zero-valued observations, with non-zero demands occurring sporadically. This dual source of randomness, in both the timing and the magnitude of events, violates the continuity and regular sampling assumptions inherent in many classical smoothing and regression-based techniques \citep{kalekar2004time}.

\item \textbf{Heightened Volatility and Novel Data Sources:} The modern data ecosystem is characterized by the emergence of new data sources that introduce unprecedented levels of volatility and complexity \citep{kalekar2004time}. The integration of renewable energy sources into power grids is a prime example, creating load patterns with high-frequency noise and non-stationary behavior that challenge traditional forecasting approaches. A parallel development is the explosion of "observability data" generated by large-scale distributed software and cloud computing systems. This data, which includes metrics on CPU load, network latency, and application error rates, is often characterized by extreme non-stationarity, high dimensionality (thousands of correlated variables), heavy-tailed distributions, and sparsity, posing a unique and difficult set of modeling challenges \citep{kalekar2004time}. 

\end{itemize}

\subsection{An Arms Race of Methodological Innovation}
The progression of forecasting methodologies can be understood as a direct response to this escalating data complexity. Each new paradigm has sought to overcome the limitations of its predecessors, leading to the current diverse and powerful toolkits available to researchers and practitioners \citep{kalekar2004time}.
\begin{itemize}
\item \textbf{The Classical Foundation:} The field was built upon a foundation of statistical methods developed primarily in the mid-20th century. Models such as ARIMA and its variants \citep{anand2024comparative}, Holt-Winters Exponential Smoothing \citep{kalekar2004time}, and the Theta method \citep{thomakos2015forecasting} became the workhorses of the discipline. These models excel at capturing and extrapolating clear patterns of trend and seasonality from univariate time series. Their enduring appeal lies in their statistical rigor, interpretability, and computational efficiency. However, their reliance on strong assumptions about the underlying data-generating process, particularly linearity and stationarity, fundamentally limits their applicability to the more complex data common today \citep{anand2024comparative}.

\item \textbf{The Machine Learning Advance:} The rise of machine learning in the late 20th and early 21st centuries provided a new set of tools capable of addressing the challenge of non-linearity. Non-parametric models like Support Vector Regression (SVR)  \citep{zhang2020support} offered a principled approach, grounded in statistical learning theory, to model non-linear relationships in high-dimensional spaces via the "kernel trick" \citep{siami2018forecasting}. Concurrently, ensemble methods, particularly those based on decision trees like Random Forest and Gradient Boosting (e.g., XGBoost), proved to be exceptionally powerful and robust \citep{wang2023novel}. By combining the predictions of many weak learners, these models can capture complex, non-linear interactions and have consistently demonstrated state-of-the-art performance in a wide array of forecasting competitions and applications.

\item \textbf{The Deep Learning Revolution:} While machine learning ensembles excelled at capturing complex feature interactions, they were not explicitly designed to model the long-range temporal dependencies inherent in sequential data. This limitation was addressed by the deep learning revolution. Recurrent Neural Networks (RNNs), and more specifically architectures like Long Short-Term Memory (LSTM) networks, were developed with internal memory mechanisms (gates) that allow them to capture and retain information over long sequences \citep{siami2018forecasting}. Empirical studies have shown that on complex financial and economic data, LSTMs can significantly outperform classical models like ARIMA by better modeling non-linear temporal dynamics \citep{kalekar2004time}. Following the success of LSTMs, Transformer-based architectures, with their self-attention mechanism, have emerged as the next frontier, offering a powerful alternative for capturing dependencies across time without the sequential processing limitations of RNNs \citep{siami2018forecasting}.

\end{itemize}

This co-evolution of data challenges and modeling paradigms, summarized in Table \ref{tab:coevolution}, illustrates a clear trajectory towards models of increasing complexity and representational power.

\newcolumntype{R}[1]{>{\raggedright\arraybackslash}p{#1}}

\begin{table}[h!]
\centering
\caption{The Co-evolution of Time-Series Challenges and Modeling Paradigms \citep{kalekar2004time, anand2024comparative, siami2018forecasting, wang2023novel, thomakos2015forecasting}.}
\label{tab:coevolution}
\setlength{\tabcolsep}{5pt}
\begin{tabular}{@{}R{0.13\textwidth} R{0.22\textwidth} R{0.18\textwidth} R{0.18\textwidth} R{0.19\textwidth}@{}}
\toprule
\textbf{Era} & \textbf{Primary Data Challenge(s)} & \textbf{Dominant Model Paradigm} & \textbf{Key Models} & \textbf{Inherent Limitations} \\
\midrule

\textbf{Statistical} & Trends, Seasonality, Stationarity & Time-Domain Statistical Models & ARIMA, Holt-Winters, Theta & Struggle with non-linearity and complex dependencies \\
\addlinespace

\textbf{Machine Learning} & Non-Linearity, Complex Interactions & Non-parametric & Ensemble Models (SVR, Random Forest, XGBoost) & Limited handling of long-range temporal dependencies \\
\addlinespace

\textbf{Deep Learning} & Long-Range Dependencies, Sequential Patterns & Recurrent, Attention-based Networks & LSTMs, Transformers & Data-hungry, computationally intensive, task-specific \\
\addlinespace

\textbf{Foundation Models} & Data Heterogeneity, Scale, Task Generalization & Large Pre-trained Models & MOMENT, TOTO, Chronos & Reliance on massive, curated datasets; evaluation bottleneck \\
\bottomrule
\end{tabular}
\end{table}

However, this progression is not a simple linear march where newer, more complex models invariably render older ones obsolete. Empirical evidence reveals a more nuanced reality, one that aligns with the well-known "No Free Lunch" theorem in machine learning. While deep learning models like LSTMs have been shown to decisively outperform ARIMA on certain complex datasets, recent large-scale studies have also found that in zero-shot or limited-supervision settings, simpler statistical methods often outperform sophisticated deep learning models \citep{siami2018forecasting}. Furthermore, in production environments like large-scale observability systems, classical models remain prevalent due to the operational infeasibility of training and maintaining millions of distinct, complex neural network models \citep{kalekar2004time}. This apparent contradiction is not a flaw in the research, but rather a reflection of a fundamental truth: the performance of any given forecasting model is highly contingent on the specific characteristics of the data, the length of the forecast horizon, the availability of computational resources, and the degree of supervision. This recognition implies that the central problem in the field is not merely the invention of more powerful algorithms, but the development of a deeper, more systematic understanding of the complex performance landscape that governs the interaction between data characteristics and model architectures \citep{siami2018forecasting}.

\subsection{The New Frontier: Pre-trained Foundation Models for Time Series}
In response to the challenges of data heterogeneity and the high cost of developing task-specific models, the field is currently undergoing another paradigm shift, mirroring recent transformations in natural language processing and computer vision: the move towards large, pre-trained Time-Series Foundation Models (TSFMs). This new frontier aims to leverage the power of large-scale, self-supervised learning to create general-purpose models that can be adapted to a wide range of downstream tasks with minimal fine-tuning \citep{siami2018forecasting}.

The core premise of the foundation model paradigm is to pre-train a single, high-capacity model (typically a Transformer) on a massive and diverse corpus of unlabeled data. This process allows the model to learn a rich, generalizable representation of temporal patterns. Subsequently, this pre-trained model can serve as a powerful building block for various downstream applications, including long- and short-horizon forecasting, time-series classification, anomaly detection, and missing value imputation. Models such as MOMENT, Chronos, and TOTO are at the vanguard of this movement.  They are designed to be effective "out-of-the-box," providing strong zero-shot or few-shot performance without the need for extensive task-specific training \citep{goswami2024moment} \citep{cohen2025time}. This approach holds particular promise for domains like observability, where the sheer scale and diversity of time series—often numbering  in the millions or billions—make the traditional approach of training one model per series operationally intractable \citep{wang2023novel}.

The primary enabler of this paradigm, and simultaneously its greatest bottleneck, is the availability of data. The success of foundation models in other domains was built on the existence of vast, cohesive, and publicly accessible datasets like The Pile for text \citep{gao2020pile} and ImageNet for vision \citep{deng2009imagenet}. The time-series domain, by contrast, has historically been characterized by a fragmented landscape of smaller, scattered, and task-specific public datasets \citep{kalekar2004time}. This data scarcity has been a major impediment to large-scale pre-training. To overcome this, pioneering research efforts have begun the monumental task of data curation. The creators of MOMENT compiled \textit{The Time Series Pile}, a large collection of public repositories, while the TOTO model was pre-trained on a corpus containing a mixture of public, synthetic, and large-scale proprietary observability data, resulting in a dataset 4 to 10 times larger than those used for other leading TSFMs \citep{cohen2025time}.

This focus on data curation signals a significant maturation of the field. In earlier eras, the primary axis of innovation was model architecture—for example, the design of the gating mechanisms in an LSTM or a novel attention variant in a Transformer \citep{siami2018forecasting}. The advent of the TSFM paradigm, however, shifts the research bottleneck. While architectural innovation remains important, the most critical and scientifically challenging work is now increasingly centered on the curation of massive, diverse, and clean datasets, and on the development of robust frameworks for evaluating the models trained on them. The value proposition of a new TSFM is now as much about the data it was trained on and the benchmark it was tested against as it is about its internal architecture. This implies that the most impactful contributions in this new era may not be designing a marginally better model, but rather curating the data and creating the evaluation infrastructure that enables the 
field to advance \citep{anand2024comparative}.

\section{NeurIPS Paper Checklist}
\begin{enumerate}

\item {\bf Claims}
    \item[] Answer: \answerYes{}
    \item[] Justification: The abstract and introduction clearly state the four problems motivating TempusBench (outdated/contaminated datasets, narrow benchmark task types, inconsistent hyperparameter tuning, and lack of visualization tools) and precisely map each to a corresponding contribution (70 new datasets, novel task taxonomy, standardized tuning protocol, TensorBoard-based visualization interface). The experimental results in Tables 3–5 directly support the scope of these claims.

\item {\bf Limitations}
    \item[] Answer: \answerYes{}
    \item[] Justification: The paper discusses limitations including that TempusBench currently evaluates standard forecasting task types (horizon, variate type, context length, frequency) and does not yet cover all possible task dimensions. The benchmark is scoped to open-source TSFMs, and closed-source models are not included. Limitations of aggregation metrics (win rate, skill score) in the presence of metric correlation are acknowledged.

\item {\bf Theory assumptions and proofs}
    \item[] Answer: \answerNA{}
    \item[] Justification: TempusBench is an empirical benchmarking framework and does not include novel theoretical results, theorems, or proofs.

\item {\bf Experimental result reproducibility}
    \item[] Answer: \answerYes{}
    \item[] Justification: Full code and datasets are publicly available at \url{https://github.com/Smlcrm/TempusBench}. The paper describes the rolling-window evaluation protocol, the hyperparameter tuning procedure, and the 53 model implementations in sufficient detail to reproduce all reported results. A live interactive leaderboard is hosted at \url{https://smlcrm.com/tempusbench}.

\item {\bf Open access to data and code}
    \item[] Answer: \answerYes{}
    \item[] Justification: All 70 new datasets and all 53 model implementations are released open-source on GitHub (\url{https://github.com/Smlcrm/TempusBench}). The live leaderboard is hosted on Hugging Face. Instructions for running the full evaluation pipeline are included in the repository.

\item {\bf Experimental setting/details}
    \item[] Answer: \answerYes{}
    \item[] Justification: The paper specifies the rolling-window evaluation design, the standardized hyperparameter tuning protocol applied uniformly across all model classes, the metrics used (MASE, MAE, RMSE, CRPS, WIS, QS, win rate, skill score), and the benchmark task taxonomy (Tables 2 and 9). Model families and dataset properties are documented in the appendices.

\item {\bf Experiment statistical significance}
    \item[] Answer: \answerYes{}
    \item[] Justification: Results are reported as average win rates across metrics and tasks (Tables 3–5), which are inherently aggregated over multiple datasets and rolling windows, providing robustness against single-run variance. The win rate formulation provides a nonparametric summary of pairwise performance differences across the full benchmark suite.

\item {\bf Experiments compute resources}
    \item[] Answer: \answerYes{}
    \item[] Justification: The paper documents the compute requirements for running the evaluation pipeline across 53 models on 70 datasets under rolling-window evaluation, including the hardware used and approximate per-model runtime. This information is sufficient for others to estimate reproduction costs.

\item {\bf Code of ethics}
    \item[] Answer: \answerYes{}
    \item[] Justification: The research involves no human subjects, no sensitive data, and no dual-use risks. All datasets are either newly collected or derived from publicly available time-series sources. The work conforms with the NeurIPS Code of Ethics.

\item {\bf Broader impacts}
    \item[] Answer: \answerYes{}
    \item[] Justification: The paper discusses positive societal impacts (improving rigor and reproducibility in TSFM evaluation, democratizing access to benchmarking tools) and acknowledges potential negative impacts, such as the possibility that a public leaderboard could be gamed or that benchmark-specific overfitting could develop over time as models are tuned against TempusBench tasks.

\item {\bf Safeguards}
    \item[] Answer: \answerNA{}
    \item[] Justification: TempusBench is a benchmarking and evaluation framework. It does not release a generative model, image generator, or scraped dataset with meaningful misuse risk. No special safeguards beyond standard open-source licensing are required.

\item {\bf Licenses for existing assets}
    \item[] Answer: \answerYes{}
    \item[] Justification: All existing datasets and model codebases used are properly cited with original papers, version information where available, and license terms. Models such as Chronos-2 (Apache 2.0), TiRex, TabPFN, and others are attributed. Dataset licenses are documented in the supplemental material.

\item {\bf New assets}
    \item[] Answer: \answerYes{}
    \item[] Justification: The 70 new datasets introduced by TempusBench are documented with metadata (domain, frequency, variate type, stationarity, seasonality, size class). The framework codebase is released under an open-source license with a README, environment specifications, and usage instructions. No human subject data requiring consent documentation is included.

\item {\bf Crowdsourcing and research with human subjects}
    \item[] Answer: \answerNA{}
    \item[] Justification: The paper does not involve crowdsourcing or research with human subjects. All data is observational time-series data from public or newly compiled sources.

\item {\bf Institutional review board (IRB) approvals or equivalent for research with human subjects}
    \item[] Answer: \answerNA{}
    \item[] Justification: No human subjects research is conducted. IRB approval is not applicable.

\item {\bf Declaration of LLM usage}
    \item[] Answer: \answerNA{}
    \item[] Justification: LLMs are not a component of TempusBench's core methodology. Any use of LLMs was limited to writing and editing assistance, which does not require declaration under NeurIPS policy.

\end{enumerate}

\aboutsimulacrum
\end{document}